%% file: main.tex
\documentclass[sigconf]{acmart}

\copyrightyear{2022} 
\acmYear{2022} 
\setcopyright{acmcopyright}\acmConference[ICSE '22]{44th International Conference on Software Engineering}{May 21--29, 2022}{Pittsburgh, PA, USA}
\acmBooktitle{44th International Conference on Software Engineering (ICSE '22), May 21--29, 2022, Pittsburgh, PA, USA}
\acmPrice{15.00}
\acmDOI{10.1145/3510003.3510088}
\acmISBN{978-1-4503-9221-1/22/05}

\newcommand{\name}{\texttt{EREBA}\xspace} 
\usepackage{amsmath}
\usepackage{slashbox}
\usepackage[linesnumbered,boxed,ruled,nofillcomment]{algorithm2e}
\usepackage{algpseudocode}
\usepackage{adjustbox}
\usepackage{mdwlist}
\usepackage{graphicx} 
\usepackage{verbatim}
\usepackage{verbatimbox}
\usepackage{fancyvrb}
\usepackage{MnSymbol}
\usepackage{multirow}
\usepackage{slashbox}
\usepackage{balance}
\usepackage{bold-extra}
\usepackage{caption}
\usepackage{subcaption}
\usepackage{diagbox}
\usepackage{mathtools} 
\usepackage{numprint}
\usepackage{tabularx}

\newcolumntype{Y}{>{\centering\arraybackslash}X}
\renewcommand{\arraystretch}{2}

\def\tool{\texttt{EREBA}\xspace}

\begin{document}

\title{EREBA: Black-box Energy Testing of Adaptive Neural Networks} 

\author{Mirazul Haque}

\authornote{Both authors contributed equally to this research.}
\email{mirazul.haque@utdallas.edu}
\affiliation{%
  \institution{The University of Texas at Dallas}
}
\author{Yaswanth Yadlapalli}
\email{yaswanth.yadlapalli@utdallas.edu}
\authornotemark[1]
\affiliation{%
  \institution{The University of Texas at Dallas}
}

\author{Wei Yang}
\email{wei.yang@utdallas.edu}
\affiliation{%
  \institution{The University of Texas at Dallas}
}

\author{Cong Liu}
\email{cong@utdallas.edu}
\affiliation{%
  \institution{The University of Texas at Dallas}
}



\input{abstract}
\maketitle
\input{newintro}
\input{definition.tex}
\input{preliminary_results}

\input{Approach}
\input{new_eval.tex}

\input{discussion.tex}

\input{related}
\input{CONCLUSION}

\newpage
\bibliographystyle{ACM-Reference-Format}
\bibliography{main.bib}

\end{document}

%% file: abstract.tex
\begin{abstract}


Recently, various Deep Neural Network (DNN) models have been proposed for environments like embedded systems with stringent energy constraints. The fundamental problem of determining the robustness of a DNN with respect to its energy consumption (energy robustness) is relatively unexplored compared to accuracy-based robustness. This work investigates the energy robustness of Adaptive Neural Networks (AdNNs), a type of energy-saving DNNs proposed for many energy-sensitive domains and have recently gained traction. We propose EREBA, the first black-box testing method for determining the energy robustness of an AdNN. EREBA explores and infers the relationship between inputs and the energy consumption of AdNNs to generate energy surging samples. Extensive implementation and evaluation using three state-of-the-art AdNNs demonstrate that test inputs generated by EREBA could degrade the performance of the system substantially.
The test inputs generated by EREBA can increase the energy consumption of AdNNs by 2,000\% compared to the original inputs. Our results also show that test inputs generated via EREBA are valuable in detecting energy surging inputs.

\end{abstract}

%% file: newintro.tex
\section{Introduction} 
 



Recently there has been a considerable amount of research in developing energy-saving DNN models to allow state-of-art DNNs with high computational costs to be deployed in mobile and embedded architecture. \textit{Adaptive Neural Networks (AdNNs)}~\cite{figurnov2017spatially,teerapittayanon2016branchynet,bolukbasi2017adaptive} are energy-saving DNN models that determine when to switch off certain parts of the network to reduce the number of computations.

Because an AdNN model determines which parts of the neural network to run based on inputs, an adversary's ability to surge the energy consumption by carefully crafting inputs is a crucial concern in energy-critical environments. For example, AdNNs like BlockDrop~\cite{wu2018blockdrop} and SkipNet \cite{wang2018skipnet}  can reduce the computations in ResNet significantly and an alteration on the input can nullify a large portion of the reduced computations, invalidating the models' purpose. Such behavior would lead the app or software using an AdNN model to consume energy erratically, resulting in devices' power failure and disastrous consequences. Thus, there is a strong need to provide a systematic testing method to find energy hotspots in the model and filter out potential ``power-surging'' inputs that will negatively impact the model's performance.

Creating testing inputs to increase  the energy consumption of a DNN model is challenging because inferring the relation between energy consumption and input is a challenging task. Unlike inferring the relation between input and output, where we can find the derivatives from a series of computation functions in the model, energy consumption can only be measured by running the model. Traditional DNN testing methods~\cite{pei2017deepxplore,xie2019deephunter,ma2018deepgauge,tian2018deeptest} and traditional adversarial attacks~\cite{papernot2016limitations,carlini2017towards,goodfellow2014explaining} on DNNs have been designed to create carefully crafted synthetic testing inputs using the gradient of generated output with respect to the input. 
However, for energy testing, it is unclear whether a change in the input induces an increase or decrease in the energy consumption of the model. To the best of our knowledge, ILFO~\cite{MirazILFO} is the first work that seeks to formulate all types of AdNN's energy robustness (Section \ref{sec:def}) problem by modeling the relation between input and intermediate output~\cite{MirazILFO} (DeepSloth~\cite{hong2020panda} only evaluates energy robustness of Early-termination AdNNs). 

However, our investigations (Section \ref{sec:transfer}) show that ILFO generated energy surging samples lack traditional transferability, \textit{i.e.,}, the adversarial samples generated by ILFO for a target AdNN cannot be applied to a new AdNN to increase its energy consumption. Therefore, the traditional black-box accuracy testing method of DNNs using surrogate model \cite{papernot2017practical,cheng2019improving,liu2016delving} can not be used for energy robustness evaluation. 
Therefore, ILFO generated samples cannot evaluate the energy robustness of AdNNs in a black-box scenario.

This paper presents \tool (\textbf{E}nergy \textbf{R}obustness using \textbf{E}stimator \textbf{B}ased \textbf{A}pproach) to perform energy testing on AdNNs under the black-box setting where there is no prior knowledge known about the AdNN model.
To our knowledge, this is the first attempt in this direction.
\tool aims to evaluate the energy robustness of AdNN and identify inputs that will negatively impact the model’s performance.
Specifically, we develop two testing methods to assess any given AdNN model's energy robustness, namely Input-based testing and Universal testing.
Input-based testing evaluates energy robustness where testing inputs are semantically meaningful to the AdNN (\textit{e.g.,} meaningful images, compilable programs). On the other hand, universal testing evaluates worst-case energy robustness where each testing input maximizes the energy consumption for each target AdNN.


For generating testing inputs for AdNNs in a black-box setting, it is needed to find a relation between input and energy consumption of AdNNs. Based on the working mechanism of AdNNs, we know that different numbers of residual blocks/layers are activated during inference for different inputs. The number of activated blocks/layers during inference has a semi-linear (step-wise) relation with energy consumption, which can also be noticed in Figure~\ref{fig:dicussion}. Through this step-wise relation between the number of activated blocks and energy consumption, we can conclude that input and energy consumption of AdNNs are related. Because of this reason, \name is able to learn a decent approximation of the energy consumption of an AdNN given the input. 
Based on such approximation, \name then generates input perturbations that significantly increase the energy consumption of the AdNN.

We evaluate \name{} on four criteria: effectiveness, sensitivity, quality, and robustness using the CIFAR-10 and CIFAR-100 datasets \cite{krizhevsky2009learning,cifar10,cifar100}. First, to evaluate the effectiveness of the testing inputs generated by EREBA, we calculate the energy required for AdNNs to classify these inputs while running on an Nvidia TX2 server.  We then compare this value with the energy required by the inputs generated from common corruptions and perturbations techniques ~\cite{hendrycks2019robustness} and a surrogate model-based approach. We observe that \name{} is twice as effective. The sensitivity of \name{} is measured through the behavior of the energy consumption of testing inputs generated while limiting the magnitude of perturbation allowed, which enables a comparison between the AdNN models' energy robustness. The quality of the generated testing inputs is evaluated against the original input through Peak Signal-to-Noise Ratio (PSNR) and Structural Similarity Index (SSIM) ~\cite{psnr,SSIM}. Finally, the robustness of \name{} is demonstrated by providing corrupted input images for the generation of testing inputs, which reveals the capability of the estimator model to imitate the shortcomings of the target AdNN.
We further demonstrate two ways to show how \name{} generated test inputs can help to increase energy robustness: through input filtering and gradient-based detection.


Our paper makes the following contributions:

\begin{itemize}
    \item An approach, \name, the first energy-oriented black-box testing methodology for AdNNs.
    \item A systematic empirical study on transferability of energy-based testing inputs.
    \item Four evaluations to demonstrate the effectiveness, sensitivity, quality, and robustness of \name.
    \item Two applications demonstrating the energy-saving capability  of \name.
\end{itemize}

%% file: definition.tex
\section{Background}
\label{sec:background}

\subsection{Energy Robustness}
\label{sec:def}
ILFO \cite{MirazILFO} has defined the energy robustness of a DNN as the stability of the model's energy consumption after getting a perturbed input.
However, a model's energy robustness should not only depend on the inputs that belong to the training data distribution of the model. Energy robustness should also be evaluated based on the out-of-distribution inputs.
Because of this reason, we define two types of energy robustness for DNNs: Input-based Energy Robustness ($E_i$) and Universal Energy Robustness ($E_u$). 

$E_i$ is defined by the maximum energy consumed by the model for an input which belongs to the training data distribution of the model. Let us assume, $x$ is an input that is within the data distribution of a DNN $f$. We want to add perturbation $\delta$ to $x$ such that energy consumption is maximum. In that scenario, $E_i$ can be represented as,

$$  {E_i} = -\underset{\delta \in R}{\textrm{max}} \; (ENG_f(x + \delta)-ENG_f(x))$$, where $R$ is set of admissible perturbations such that $x + \delta$ remains within distribution, and $ENG_f$ represents the energy consumption of DNN $f$.

$E_u$ can be described as the highest possible energy consumed by a model for any input. Inputs used to measure $E_u$ can be out-of-distribution inputs also. For a DNN $f$ and any input $x$, $E_u$ can be represented as,

$$  {E_u} = -\underset{x}{\textrm{max}} \; ENG_f(x)$$, where $ENG_f$ represents energy consumption of DNN. By increasing the value of $E_i$ and $E_u$, energy robustness of a model can be increased.

\subsection{AdNNs}
\label{sec:adnn}
The main objective of AdNNs is to minimize executing layers in a Neural Network while maintaining reasonable accuracy. The AdNNs can be divided mainly into two types: \textit{Conditional-skipping AdNNs} \cite{wu2018blockdrop,wang2018skipnet} and \textit{Early-termination AdNNs} \cite{bolukbasi2017adaptive,teerapittayanon2016branchynet}. Both types of AdNNs reduce computations if their intermediate output values satisfy predefined conditions. For reducing computations, Conditional-skipping AdNNs skip a few layers or residual blocks \footnote{Residual block consists of multiple layers whose output is determined by adding the output of the last layer and input to the block.} (in the case of ResNet), while Early-termination AdNNs terminates the operations within a block or network early.

%% file: preliminary_results.tex
\section{Transferability of Energy-based Testing Inputs.}
\label{sec:transfer}
In this section, by carrying out a preliminary study, we show that traditional transferability does not exist in energy testing inputs, and existing technique like attacking surrogate model to generate accuracy-based testing inputs cannot be applied in energy testing. For traditional accuracy-based testing, transferability refers to the property that adversarial examples generated for one model may also be misclassified by another model.

\noindent \textbf{Motivation.} In a black-box setting, existing techniques~\cite{papernot2017practical,cheng2019improving,liu2016delving} evaluate the accuracy-robustness of DNNs based on the traditional transferability of adversarial samples~\cite{papernot2016transferability}
. 
Adversarial examples of DNNs are perturbed inputs close to the original correctly classified inputs but are misclassified by DNNs.
Because adversarial examples are commonly used as testing inputs to measure the robustness of the neural networks~\cite{ma2018deepmutation, xie2019deephunter}),
we will also use the term\textit{ testing inputs} to refer to the adversarial examples in this paper.
Goodfellow \textit{et al.} and Szegedy \textit{et al.} \cite{goodfellow2014explaining,szegedy2013intriguing}  have concluded that accuracy-based testing inputs on a traditional DNN model are transferable.  Therefore, adversarial examples generated by attacking a surrogate DNN model can be applied to other DNNs for evaluating robustness.
In this section, we investigate if traditional transferability, which is used for measuring accuracy robustness in a black-box setting, can be applicable for energy-based testing inputs.

\begin{table}[h]
\centering
\setlength{\tabcolsep}{0.3em}
{\renewcommand{\arraystretch}{1}
\scriptsize
\caption{$ITP$ among different architectures.  RN is ResNet, BD is BlockDrop, BN is BranchyNet. BM represents Base Model, while TM is Target Model.}
    \label{tab:transfer}
 \begin{tabular}{|l| r| r| r| r|} 
 \hline
 \diagbox[width=7em]{BM}{TM} & RAN & BD (RN 110)& BN& BD (RN 32) \\ [0.5ex] 
 \hline
 RAN &100.0 &46.0 & 41.0&12.5\\
 \hline
 BD (RN 110) &  64.0 & 100.0& 68.0&72.4\\
 \hline
 BN & 61.0 & 52.0& 100.0&4.0\\  
 \hline
 
 BD (RN 32) &  5.5 & 45.0& 75.2&100.0\\
 \hline
\end{tabular}}

\end{table}

\begin{table}[h]
\centering
\setlength{\tabcolsep}{0.3em}
{\renewcommand{\arraystretch}{1}
\scriptsize
\caption{$ETP$ among different architectures. RN is ResNet, BD is BlockDrop, BN is BranchyNet. BM represents Base Model, while TM is Target Model.}
    \label{tab:transfer2}
 \begin{tabular}{|l| r| r| r| r|} 
 \hline
 \diagbox[width=7em]{BM}{TM} & RAN & BD (RN 110)& BN& BD(RN 32) \\ [0.5ex] 
 \hline
 RAN &100.0 &0.1 & 40.0&-1.8\\
 \hline
 BD (RN 110) &   200.0 & 100.0& 350.0&52.5\\
 \hline
 BN & 38.0 & 3.0& 100.0&-3.8\\  
 \hline
 
 BD (RN 32) &  -3.5 & 10.0& 228.0&100.0\\
 \hline
\end{tabular}}

\end{table}

\noindent \textbf{Preliminary study.} We have conducted a study to investigate the traditional transferability of energy-based testing input on AdNNs. To our knowledge, this is the first effort to explore the transferability of energy-based testing input on AdNNs. We define base and target models for this study. The white-box attack is performed on the base model, and the target model classifies the testing input.
We focus on two metrics to measure transferability: the percentage of the transferable adversarial inputs and the average percentage of the transferable energy consumption increase. We define two terms: \textit{Effectiveness Transferability Percentage} \textit{(ETP)} and \textit{Input Transferability Percentage} \textit{(ITP)}. \textit{ETP} is defined based on \textit{IncRF}, which is the fractional increase in AdNN-reduced floating-point operations (FLOPs) after feeding energy-based testing inputs. We also define $P_b$ and $P_t$, the average \textit{IncRF} on base and target models, respectively, with the same testing inputs. We define \textit{ETP}=$(P_t / P_b) \times 100$. \textit{ITP} is defined as the percentage of testing inputs for which the FLOPs count during inference increases in the target model. For an attack, if \textit{ITP} is high, it means that most of the generated testing inputs for the base model can also increase the energy consumption in the target model. If \textit{ETP} is high, it means that the average increase in the target model's energy consumption is comparable with the base model. Thus, if both \textit{ETP} and \textit{ITP} are high, then it confirms transferability in the attack.



For example, we attack the base model and perturb ten inputs. If the average $IncRF$ on base model is 0.5, \textit{i.e.}, $P_b=0.5$. If seven out of ten testing inputs increase the FLOPs on the target model, $ITP$ will be 70 \%. For target model, The average \textit{IncRF} is 0.3. $P_t$ would be $0.3$ and \textit{ETP}=$(0.3 / 0.5) \times 100=60 \%$. We have set multiple \textit{thresholds} of $ITP$ and $ETP$ to determine whether an attack is transferable.
In this study, we explore how many combinations of base AdNN and target AdNN exceeds the different \textit{ITP}  and \textit{ETP} thresholds.

We have conducted a transferability study on four AdNN models: RANet \cite{yang2020resolution}, BlockDrop (ResNet-38, ResNet-110) \cite{wu2018blockdrop}, and BranchyNet \cite{teerapittayanon2016branchynet}. 1000 images sampled from the CIFAR-10 dataset were used in this study.
We generate testing inputs using the ILFO attack~\cite{MirazILFO} (Figure \ref{fig:approach1} Sub-figure I) on the base model.
Tables \ref{tab:transfer} and \ref{tab:transfer2} show the \textit{ITP} and \textit{ETP} among the architectures. From the tables, we can see that only four combinations (out of 12) of base and target models exceed the 50\% threshold for both \textit{ITP} and \textit{ETP} (For other thresholds, we have added a table in the website). From these results, we can conclude that for the majority of AdNN models, the white-box attack is non-transferable.


Although traditional transferability may not be feasible for attacking AdNNs, the observation of this failure motivates us to develop an effective alternative. Specifically, our key observation on the non-transferability of energy-based attacks is that the energy-saving mechanisms of AdNN models behave differently for the same input, \textit{i.e.,} an input causing high energy consumption on one model might consume low energy for another model. Hence, extending any white-box attack method such as ILFO through Surrogate models is not viable in the black-box scenario.

%% file: Approach.tex
\section{Approach}
\label{sec:approach}



As we have observed that Surrogate models are not feasible to test the energy robustness of AdNNs, we develop EREBA, an Estimator model-based black-box testing system. EREBA contains two major components, namely an Estimator model and a testing input generator. Figure \ref{fig:approach1} illustrates the Estimator model’s training for an AdNN using its energy consumption on an Nvidia TX2 server. The Estimator model addresses the challenge of non-transferability discovered in the previous section. Additionally, the testing input generator in EREBA has two modes of testing: Input-based, where an input image is perturbed to achieve higher energy consumption, and Universal, where a noisy testing input that can maximize energy consumption is generated. These modes enable EREBA to assess the energy robustness of each AdNN effectively.
Figure \ref{fig:approach1} (Sub-figure ii) shows the generation of testing inputs using the trained Estimator model.



\subsection{Estimator Model Design}
\label{sec:estimator}

\begin{figure*}[t]%
	\centering
	\includegraphics[width=1\textwidth]{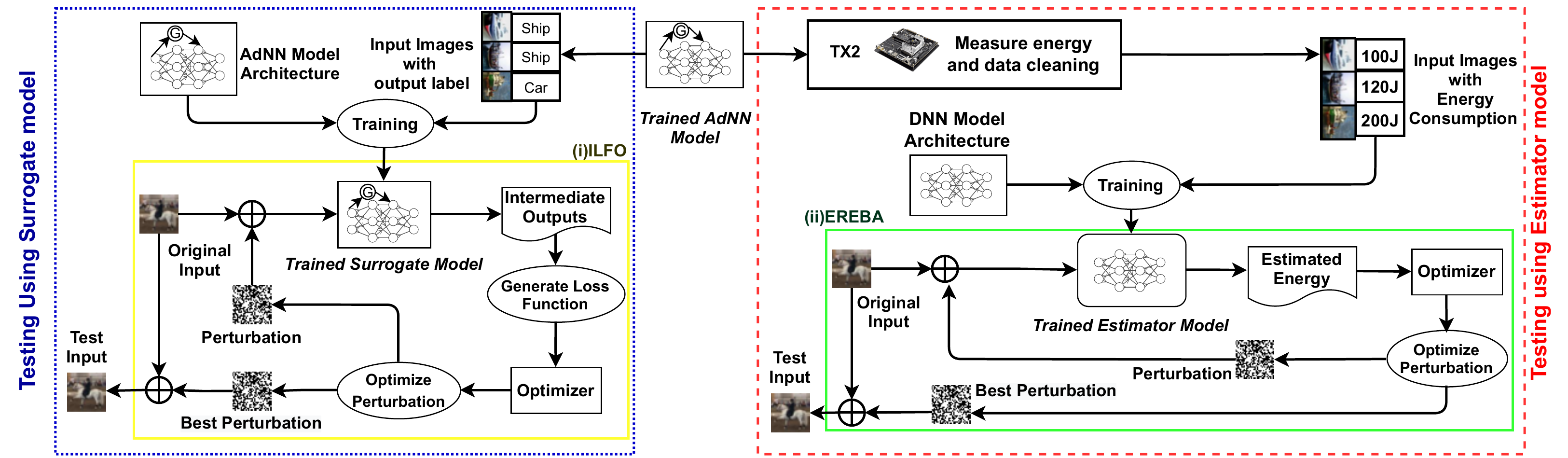}

	\caption{Difference between testing using Estimator model and Surrogate model}
	
	\label{fig:approach1}
\end{figure*}


Traditionally, misclassification black-box adversarial methods on DNNs are achieved through Surrogate models (also DNNs) trained using the output labels produced by feeding the target DNN with original images. Testing inputs generated by a white-box method against the trained Surrogate model are then used against the target DNN as illustrated in Figure 1 (inside the blue-dotted box). However, this approach is not feasible for the current use-case due to a lack of traditional transferability in white-box attacks (Section \ref{sec:transfer}). Because building such a Surrogate model with the target function of mapping an image to a class would not transfer similar energy characteristics to the Surrogate model.
The target function should be the energy-saving mechanism in an AdNN, but the output dimensions of these change with different AdNNs. So, we need a separate Surrogate model for each AdNN; this is not viable for two reasons.
First, such a model would require a new neural network architecture for each AdNN, which makes it hard to apply to future AdNN models. Second, the energy-saving mechanisms' outputs are intermediate values within an AdNN, which are not accessible in a black-box scenario.

To tackle this, we build an Estimator model to emulate the characteristics of the energy-saving mechanism in AdNNs. The feasibility of the Estimator model follows from the following two key observations.  1) We can perceive the energy-saving mechanisms' characteristics through system diagnostics such as the energy consumption for each inference, which can be observed even in a black-box setting. 2) Even though each AdNN has a different energy-saving mechanism, the resulting energy consumption is always expected to lie in a step-wise pattern \footnote{Energy consumption of processing an extra residual block or layer in an AdNN will always add similar energy consumption into the total energy consumption, making the energy consumption pattern step-wise.}; see Figure \ref{fig:dicussion} and Section \ref{sec:pattern_adnn} for more details. Thus, we seek to leverage this patterned energy consumption in our black-box approach to train an Estimator model for each target AdNN at which point the Estimator model can predict the energy consumption of each image. 

However, the energy consumption of an embedded system such as Nvidia TX2  is affected by noise from the system environment, such as background processes and dynamic frequency scaling, making it challenging to create an accurate model. Further, in a black-box environment, it is hard to categorize which data is noisy. We do not have any additional information (\textit{e.g.,} number of the executed blocks) about the inference; our approach has to tackle these challenges.

An overview of the Estimator model training is given in Figure \ref{fig:approach1} (Inside red dotted box). First, we collect energy consumption data for images used for training and,  
to address the noise in the data due to the system environment, we deactivate the dynamic frequency scaling of Nvidia TX2 and ensure that no other user processes are running. Additionally, we record the energy consumption by the target AdNN during inference of an input image twenty times and discard values, which are 50\% higher than the median value. We define the mean of the remaining values is defined as $F(x_i)$, where $x_i$ represents an input in the dataset. We define the Estimator's DNN loss function as:
    $$estimator\;loss  = \frac{1}{N} \sum_{i = 1}^{N} [F(x_i) - EST(x_i)]^2 $$
\noindent where $EST$ denote the Estimator model, and $N$ is the size of the dataset. $estimator\;loss$ is used to train the Estimator for each target AdNN, which enables the model to give a reasonable prediction of energy consumption of the target AdNN for a given input. 


	
\subsection{Testing input generator}

The objective of the testing input generator is to create testing inputs that increase the Estimator model's prediction, which in turn should increase the actual energy consumption of the AdNN. We explore two use cases of \name{}: 1) Input-based test, and 2) Universal test for measuring the Input-based and Universal energy robustness as defined in Section-\ref{sec:def}.

\subsubsection{Input-based testing}

In this use case, we modify the input image in such a way that it is imperceptible by a human, and the resulting testing input has higher energy consumption on the target AdNN. We thus add a perturbation $\delta _i$ to the input $x _i$. Picking the best $\delta _i$ ($\hat{\delta_i}$) can be formulated as:
$$  \hat{\delta_i} = \underset{\delta_i}{\textrm{argmax}} \; EST( x_i + \delta_i).$$

\noindent Additionally, we have to ensure that the magnitude of perturbation is also small as higher magnitude perturbations are more susceptible to detection. We can reformulate the maximization problem to a minimization problem as follows:
\begin{gather}\label{eq:2}
\underset{\delta_i}{\textrm{minimize}} \;  (||\delta_i|| - c \cdot EST( x_i + \delta_i)) \mbox{ where, }(x_i+\delta_i) \in [0,1]^n\, 
\end{gather}
\noindent where $c > 0$ is a hyperparameter chosen through grid search depending on the AdNN model. Also, $c$ controls the magnitude of generated perturbation ($||\delta_i||$), where a large $c$ makes the loss function more dependant on the energy estimate, allowing for larger perturbations. Whereas a smaller $c$ makes the loss function more dependant on $||\delta_i||$. Hence,  $c$ and $||{\delta_i}||$ are directly proportional. 

This constrained optimization problem in $\delta_i$ can be converted into a non-constrained optimization problem in $w_i$, where the relationship between $\delta_i$ and $w_i$ is:
    $$ \delta_i = \frac{\tanh(w_i) + 1}{2} - x_i$$
The $tanh$ function would ensure that the generated test input values stay between 0 and 1. The equivalent optimization problem in $w_i$ is:
\begin{gather}\label{eq:3}
\underset{w_i}{\textrm{minimize}} \;  \Bigg|\Bigg|\frac{\tanh(w_i) + 1}{2} - x_i\Bigg|\Bigg| - c \cdot EST\Bigg(\frac{\tanh(w_i) + 1}{2}\Bigg)
\end{gather}

\subsubsection{Universal Testing} 
\label{sec:universal_testing}

In this use case, \name{} generates a testing input only using the Estimator model. Unlike Input-based testing, which adds human imperceptible perturbation to original images, universal testing creates noisy testing inputs, which can maximize the energy consumption of the target DNN independent of the input. The intuition behind this testing is that adversaries can send noisy testing inputs exclusively to increase the system's energy consumption because human perception may not be a concern in every scenario. Hence we modify the optimization function from equation \ref{eq:3} to: 
\begin{gather}\label{eq:4}
\underset{w_i}{\textrm{minimize}} \; - EST\Bigg(\frac{\tanh(w_i) + 1}{2}\Bigg)
\end{gather}

\begin{algorithm}[h]
\scriptsize
  \SetKwInOut{Input}{Inputs}\SetKwInOut{Output}{Outputs}
  \caption{Testing input generation using \name }\label{alg:L2}
  \Input{$x_i:Input$ $Image$}
  \Output{$f_i:Perturbed$ $Image$
  }
  \BlankLine
  \Begin{
    Initialize($w_i$)\\
    $T=number\_of\_iterations$\\
    $iter\_no=0$\\
    \While{$iter\_no<T$}{
        $L = loss(x_i, w_i, c)$\\
        $L_{new},w_i=Optimizer(L,w_i)$\\
        $iter\_no++$\\
    
    }
    $f_i= \frac{\tanh w_i + 1}{2}$\\
}
\end{algorithm} 

Both minimization problems can be solved through an iterative approach given by Algorithm 1, which is also illustrated in Figure~\ref{fig:approach1} (sub-figure (ii) inside green box). The algorithm outputs the testing input $f_i$ while taking the current image $x_i$ (not required in Universal test mode) as input. $w_i$ is initialized to a random tensor (multi-dimensional array) with a size equal to the input image dimension. For each iteration, the loss function of the current mode (given in equations \ref{eq:3}, \ref{eq:4}) is computed (at line 6). This loss is back-propagated, and the optimizer takes a step in the direction of the negative gradient of the loss w.r.t $w_i$ and updates $w_i$ with its next value. Once the iteration threshold ($T$) is reached, the algorithm computes and returns the testing input $f_i$ (at Line 10). In the Universal test mode, this algorithm is repeated for $N_r$ different random initializations of $w_i$, out of which $w_i$ corresponding to the lowest loss value is used for computing $f_i$.

%% file: new_eval.tex
\section{Evaluation}
\label{sec:eval}
We evaluate the performance of \tool on three popular AdNNs, \textbf{RANet}~\cite{yang2020resolution}, \textbf{BlockDrop}~\cite{wu2018blockdrop} and \textbf{BranchyNet}~\cite{teerapittayanon2017distributed,teerapittayanon2016branchynet}, in terms of four research questions (RQs):

\textbf{RQ1: Effectiveness.} How much increase in energy consumption is achievable by the testing inputs generated by \name{}?

\textbf{RQ2: Sensitivity.} How does the energy consumption of AdNNs react to limiting the magnitude of perturbation in \name{}?

\textbf{RQ3: Quality.} What is the difference in semantic quality between original images and testing inputs generated by \name{}?

\textbf{RQ4: Robustness.} Is \name{} robust against distribution shifts?

\subsection{Experimental Setup}
\label{subsec:setup}
\textbf{Datasets.} For all AdNN models, the CIFAR-10 and CIFAR-100 datasets \cite{krizhevsky2009learning,cifar10,cifar100} have been used to train the Estimator model and generate testing inputs. Both the datasets consist of 50000 training and 10000 test images, where CIFAR-10 and CIFAR-100 have 10 and 100 class labels, respectively. By using these two datasets, we show that \name{} is useful for both easier (CIFAR-10) and more complex prediction (CIFAR-100) tasks.


\noindent\textbf{Baseline.} As there are no existing black-box energy testing frameworks, we compare our technique with two different types of baseline techniques. First, we compare our techniques with real-world corruption and perturbation techniques (like fog, frost) \cite{hendrycks2019robustness}. The datasets generated from these techniques are commonly used \cite{xie2020self,geirhos2018imagenet,ovadia2019can} to test the robustness of neural networks. Second, we use a surrogate model technique that utilizes ILFO to generate testing inputs.


Common corruption techniques \cite{hendrycks2019robustness} contain different visual corruption, which includes practical corruptions like fog, snow, frost. We use 19 different corruption types,  and for each type, five visual corruptions are created from severity level one to five, resulting in a total of 95 different visual corruptions. In the images generated by common corruption techniques, the noise present in the inputs is human perceptible. 

Common perturbation techniques \cite{hendrycks2019robustness} use 14 practical perturbation types; for each original input, 30 different images are created with different amounts of perturbation.
With perturbation,  slightly perturbed images are generated that are difficult for humans to differentiate from the original images.

Other than using common corruptions and perturbations \cite{hendrycks2019robustness}, we also use the Surrogate model-based technique as a baseline. For this approach, we create a Surrogate model for each AdNN and use the Surrogate model to generate adversarial images. As we see in Tables \ref{tab:transfer} and \ref{tab:transfer2} that adversarial inputs generated using BlockDrop as the surrogate model are more effective on other AdNNs, we add a baseline that uses BlockDrop as a Surrogate model and is referred to as SURRG in the followings sections. For each AdNN, we first classify 50000 CIFAR-10 and CIFAR-100 training data on the target AdNN and based on its outputs, and we train the Surrogate model. Then we use the ILFO attack on the Surrogate model to generate test inputs. We use Surrogate model to generate both limited perturbation (Input-based SURRG) and noisy (Universal SURRG) test samples.





\begin{table}[h]
\centering
\scriptsize
\setlength{\tabcolsep}{0.3em}
{\renewcommand{\arraystretch}{1}
\caption{Hyperparameters of \name{}}
    \label{tab:HP}
\begin{tabular}{|*{56}{c|}}
\hline
Mode & AdNN & $c$ & Learning Rate & $T$ & $N_r$ \\
\hline
\multirow{3}{*}{Input-based} 
  & BlockDrop & 1 & 0.01 & 500 & - \\
\cline{2-6} & BranchyNet & 100 & 0.01 & 500 & - \\
\cline{2-6} &RANet & 10 & 0.01 & 500 & - \\
\hline
\multirow{3}{*}{Universal} 
  & BlockDrop & - & 0.01 & 500  & 30 \\
\cline{2-6} & BranchyNet & - & 0.01 & 500   & 30 \\
\cline{2-6} &RANet & - & 0.01 & 500  & 30\\
\hline
\end{tabular}}

\end{table}

\noindent\textbf{Models.} We have selected AdNN models where the range  of maximum energy consumption during inference is large (15 J-160J). We show that \name{} can be used against AdNN models with both higher and lower number of parameters.
AdNN BlockDrop is built modifying ResNet-110 architecture and trained on CIFAR-10. While for training using CIFAR-100 dataset, BlockDrop is built modifying ResNet-32 architecture because ResNet-110 BlockDrop architecture trained on CIFAR-10 dataset has shown less adaptability.
ResNet architecture starts with a 2D convolution layer, which is followed by residual blocks.  BlockDrop selects which blocks to execute through
Policy Network for BlockDrop. Both RANet and BranchyNet are multi-exit networks, where operation can be terminated in one of the earlier exits based on the confidence score of the exit. The Estimator Model is a ResNet-110, with the output fully connected ten node layer changed to a fully connected single node. The loss function is changed to the function defined in section \ref{sec:estimator}. 
Hyperparameters chosen by the Estimator model ($c$, learning rate, number of iterations ($T$)) for each AdNN model are in Table \ref{tab:HP}, reasons for choosing these parameters are given in section \ref{sec:effectiveness}. These are parameters used for the results reported in sections \ref{sec:effectiveness} and \ref{sec:quality}, whereas section \ref{sec:robustness} reports the behaviour of \name{} when $c, T$ are changed. 

\noindent\textbf{Hardware Platform.}
We use the Nvidia Jetson TX2 board for our energy consumption measurements, which are used to train the Estimator Model. By default, TX2 has a dynamic power model, which scales the CPU and GPU frequencies based upon the current system load, which adds additional uncertainty to the energy consumption measurements. To combat this, we set the TX2 board to Max-N mode, which forces CPU and GPU clock to run at their maximum possible values, which are 2.0 GHz and 1.30 GHz, respectively.

Jetson TX2 module has two power monitor chips on board for measuring power consumption. One of the power monitors measures the power consumption of CPU, GPU, and SOC as in Fig. 8, 9 of the user manual of Nvidia TX2~\cite{tx2}. During the inference process of the AdNNs, we measure the power consumption of GPU using a monitor program. Since the monitor program only uses the CPU, the program does not affect the energy measurement. Additionally, TX2 Internal power monitors have been validated by other studies such as S. Köhler et al. \cite{kohler2020pinpoint} (see Fig 1b.), where it is shown that the power measurements using the internal monitor chips collaborate with external measurement techniques. As stated in \cite{kohler2020pinpoint}, one concern with using the internal chips is that the power consumption from the carrier board, fan, and power supply is not measured, which comes out to be around 2W. However, since our study measures the effect of various testing inputs, which cannot affect such components' behavior and both measurements (testing input, original image) ignore the consumption from these components, the conclusions drawn are valid. Further, to ensure the collected energy consumption data is correct, we run the inference twenty times and discard values outlier values that are 50\% higher than the median value. The mean of the remaining values is used as the single energy consumption value reported.

\noindent\textbf{Metrics.}
We evaluate the effectiveness and robustness of \name{} using the percentage increase in energy consumption  for each target AdNN:
$$ \frac{EnergyConsumption(f_i) - EnergyConsumption(x_i)}{EnergyConsumption(x_i)} \times 100 $$
where $x_i$ is the input image provided to \name{} and $f_i$ is the testing input generated. 
For Sensitivity (RQ2) measurements, we use the increment in energy consumption in Joules (J) to better compare sensitivity between target AdNNs and use the average squared difference in pixels values between the input image and the testing input to quantify the magnitude of the perturbation. The testing inputs' quality is measured using Peak Signal to Noise Ratio (PSNR) \cite{psnr}, and Structural Similarity Index (SSIM) \cite{SSIM} because of usage of these metrics in the industry to measure image quality.

\subsection{Experimental Results}
\subsubsection{RQ1. Effectiveness}
\label{sec:effectiveness}

\begin{figure}
    \centering
    \begin{subfigure}[b]{0.23\textwidth}
        \includegraphics[width=\textwidth]{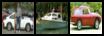}
        \caption{Original}
    \end{subfigure}
    \begin{subfigure}[b]{0.23\textwidth}
        \includegraphics[width=\textwidth]{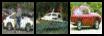}
        \caption{Testing Inputs}
    \end{subfigure}
    \caption{Testing inputs generated by \name{} for BlockDrop in Input-based testing mode}
    \label{fig:my_label}
\end{figure}

To evaluate testing effectiveness by \name{}, we have measured the average percentage increase in each AdNN model’s energy consumption between the original images in the dataset and the corresponding testing inputs generated by \name{}. We compare the effectiveness of Universal testing against the common corruption techniques \cite{hendrycks2019robustness} because, in both approaches, noise introduced to the inputs is human perceptible. Whereas Input-based testing adds imperceptible perturbation to the input, hence we compare against the common perturbation techniques  \cite{hendrycks2019robustness}. The hyperparameters chosen for each model are in Table \ref{tab:HP}.
 Parameters vary between AdNNs due to variations in input normalization, which is only applied in BranchyNet and RANet. We evaluate on images from the CIFAR-10 and CIFAR-100 datasets, which have a considerable reduction in the energy consumption on the target AdNNs ~\cite{wu2018blockdrop,yang2020resolution,teerapittayanon2016branchynet,teerapittayanon2017distributed}.  The Estimator models have been trained on 50000 CIFAR-10 and CIFAR-100 training images. We apply common corruption and perturbation techniques \cite{hendrycks2019robustness}  to CIFAR-10 and CIFAR-100 test images. As there are numerous corruption, and perturbation techniques, we only report the best performing techniques  (\textit{i.e.,} highest \textit{IncRF}) for each AdNN model (For the \textit{IncRF} values of other corruptions and perturbations, please see the website\footnote{https://sites.google.com/view/ereba/home}); Table~\ref{tab:CorrPer} reports the exact corruption/perturbation technique.

\begin{table}[h]
\centering
\scriptsize
\setlength{\tabcolsep}{0.3em}
{\renewcommand{\arraystretch}{1}
\caption{Corruptions and perturbations we have used for comparison for each model. \textit{corr} represents corruptions and \textit{per} represents perturbations.}
    \label{tab:CorrPer}
 \begin{tabular}{|l| c| c| c|} 
 \hline
 \diagbox[width=15em]{Data\\Type}{\rule{0mm}{3mm}Models} & BlockDrop & BranchyNet & RANet  \\  
 \hline
 Best Corr (CIFAR-10) & Contrast & Impulse Noise & Contrast \\ 
 \hline
 Best Per (CIFAR-10) & Gaussian Blur & Snow & Gaussian Blur \\
 \hline
 Best Corr (CIFAR-100) & Contrast & Impulse Noise & Fog \\
 \hline
 Best Corr (CIFAR-100) & Zoom Blur & Zoom Blur & Shot Noise \\
 \hline

\end{tabular}

}


\end{table}

\begin{table}[h]
\centering
\scriptsize

\setlength{\tabcolsep}{0.3em}
{\renewcommand{\arraystretch}{1}
\caption{Mean percentage increase in energy consumption of the models for CIFAR-10 dataset
against \name{} and baseline techniques.}
    \label{tab:Per}
 \begin{tabular}{|l| r| r| r|} 
 \hline
 \diagbox[width=15em]{\rule{0mm}{3mm}Perturbation \\Type}{\rule{0mm}{3mm}Models} & BlockDrop & BranchyNet & RANet  \\  [0.1ex]
 \hline
 Universal Testing (\name{}) & 97.54 & \textbf{528.51} & \textbf{1846.18} \\
 \hline
 Best Corr & 77.77 & 186.79 & 1209.00 \\
 \hline
 Universal SURRG & \textbf{137.8} & 26.87 & 302.42 \\
 \hline
 \hline
 Input-based Testing (\name{}) & 67.92 & \textbf{288.90} & 885.00 \\
 \hline
 Best Per & 16.24 & 153.72 & \textbf{1480.87} \\
 \hline
 Input-based SURRG & \textbf{135.5}& 118.39 & 554.69 \\
 \hline
 
\end{tabular}}


\end{table}

Table \ref{tab:Per} reports the mean percentage increase in energy consumption of the AdNN models under \name{} (Universal, Input-based testings) and the baselines on the CIFAR-10 dataset. Figure \ref{fig:my_label} illustrates some Input-based testing inputs generated by \name{} for BlockDrop. 
We observe that \name{} Input-based testing inputs dominate the baseline methods for mean energy increase on BranchyNet. Whereas for BlockDrop, because SURRG has BlockDrop as its architecture and ILFO is a white-box method, it is expected to outperform \name{}, a black-box method. Interestingly, for RANet, common perturbation techniques  \cite{hendrycks2019robustness} induce a much higher energy increase than the corruption techniques, which is quite different from the behavior observed in the other AdNNs. While \name{} underperforms the perturbations in terms of energy consumption increase, \name{} still outperforms SURRG. Also, we observe that for all AdNNs, samples generated through Universal testing outperform the baseline techniques.
Furthermore, for BranchyNet and RANet, due to fewer execution modes (two in BranchyNet and eight in RANet), the energy consumption is higher than BlockDrop (up to 2000 \% more than the original data) with $2^{54}$ modes.

For the CIFAR-100 dataset, Table \ref{tab:Per_c100} shows the average percentage increase in energy consumption for \name{} and the baseline techniques. We notice that \name{} generated inputs outperform common corruption and perturbation techniques for all three AdNNs. Similar to the CIFAR-10 dataset, SURRG generated inputs consume more energy than \name{} generated inputs only for the BlockDrop model. For RANet, Universal testing inputs can not significantly increase energy consumption because RANet always predicts an input with high noise as road or shrew with high confidence; therefore, the inference is stopped at initial exits, resulting in lower energy consumption. Nevertheless, the Input-based testing inputs can increase up to 4000\% energy consumption of the original inputs for the RANet model. Thus, we conclude that, on average, over all three AdNN models, \name{} performs better than any other baseline technique in terms of increasing energy consumption.
\begin{table}[h]
\centering
\scriptsize

\setlength{\tabcolsep}{0.3em}
{\renewcommand{\arraystretch}{1}
\caption{Mean percentage increase in energy consumption of the models for CIFAR-100 dataset
against \name{} and baseline techniques.}
    \label{tab:Per_c100}
 \begin{tabular}{|l| r| r| r|} 
 \hline
 \diagbox[width=15em]{\rule{0mm}{4mm}Perturbation \\Type}{\rule{0mm}{3mm}Models} & BlockDrop & BranchyNet & RANet  \\ [0.1ex] 
 \hline
 Universal Testing (\name{}) & 27.22 & \textbf{580.50} & \textbf{479.80} \\
 \hline
 Best Corr & -29.61 & 5.74 & -31.80 \\
 \hline
 Universal SURRG & \textbf{55.40} & 71.42 & -29.50 \\
 \hline
 \hline
 Input-based Testing (\name{}) & 17.27 & \textbf{283.60} & \textbf{1113.28} \\
 \hline
 Best Per & -54.28 & 37.80 & -30.73 \\
 \hline
 
 Input-based SURRG & \textbf{41.31}& 37.90 & 754.69 \\
 \hline
 
\end{tabular}}


\end{table}

\begin{figure}[!t]
    \centering
    \begin{subfigure}[t]{0.23\textwidth}
    	\includegraphics[width=\textwidth]{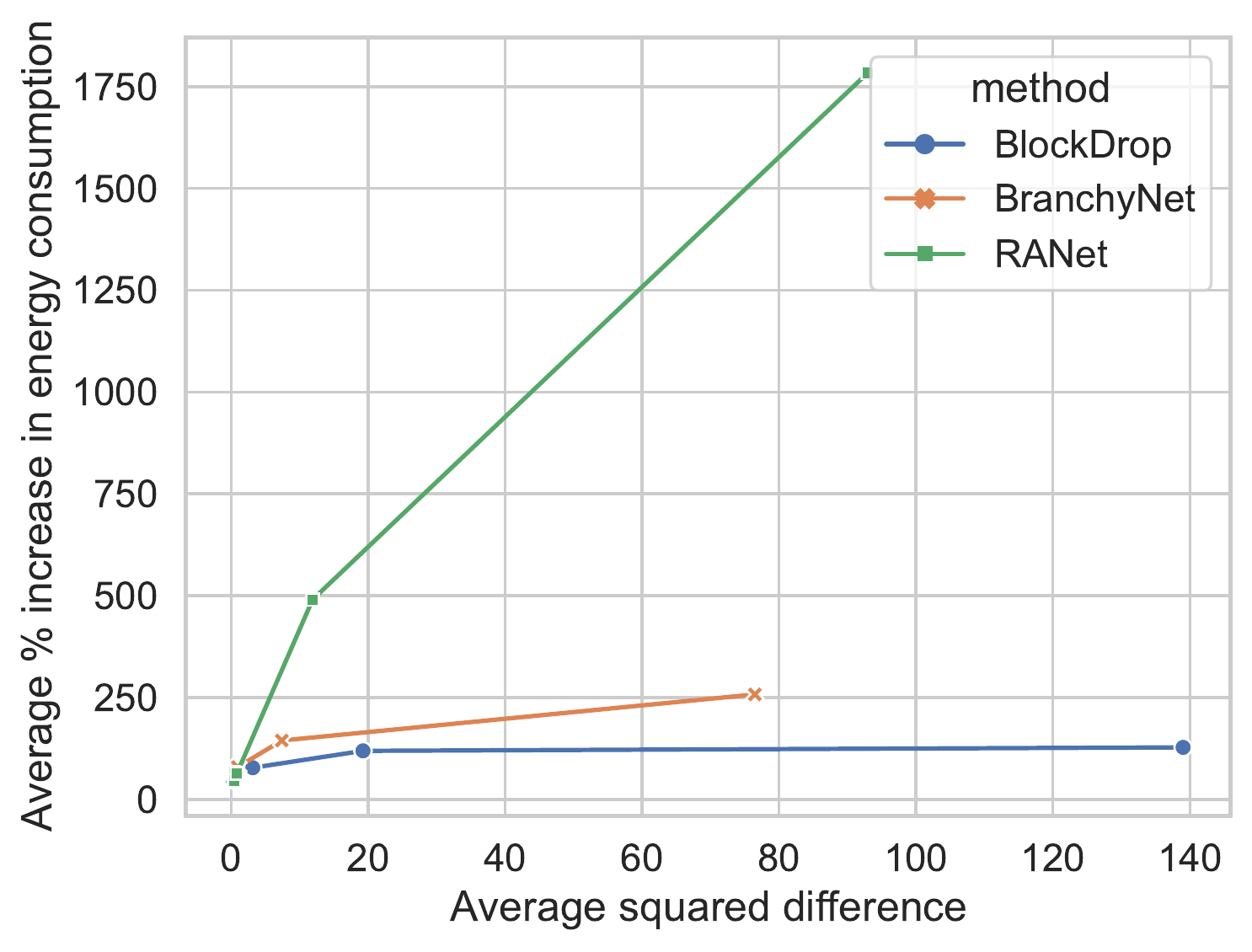}
    	\vspace{-3.8mm}
    	\caption{CIFAR-10}
    	\label{fig:robustness_cifar10}
	\end{subfigure}
    \begin{subfigure}[t]{0.23\textwidth}
        \includegraphics[width=\textwidth]{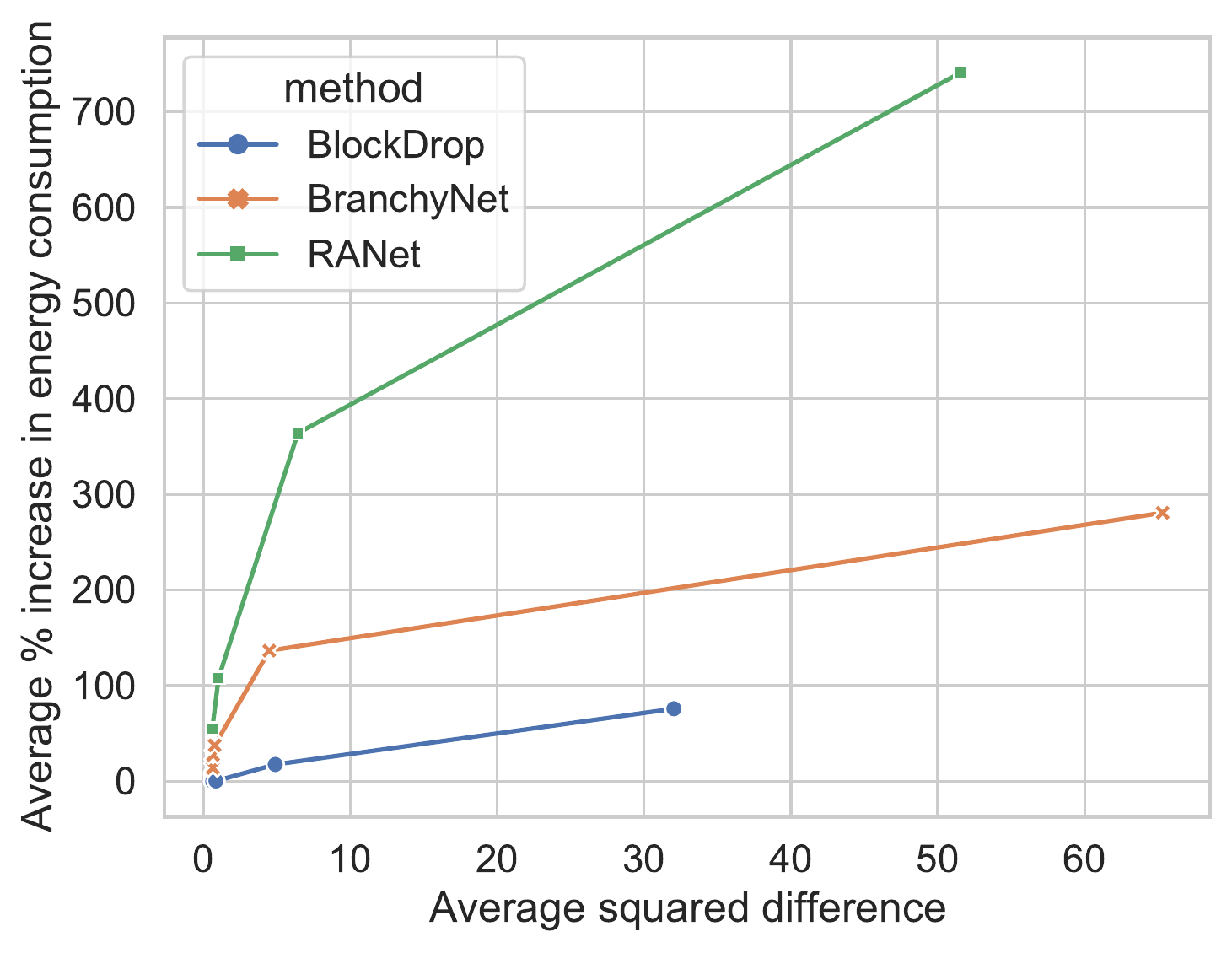}
        \caption{CIFAR-100}
        \label{fig:robustness_cifar100}
    \end{subfigure}
	
    \caption{Average energy consumption increase of testing inputs constrained by magnitude of perturbation for BlockDrop, BranchyNet and RANet.}
    \label{fig:robustness}
\end{figure}
\subsubsection{RQ2. Sensitivity}  

\label{sec:robustness}
We define the sensitivity of \name{} in terms of the magnitude of the perturbation $|\delta_i|$. Intuitively, if an AdNN model’s energy consumption spikes up with a relatively lower average perturbation magnitude, then that model is less robust.
We know that $c$ and $|\delta_i|$ are directly propositional, given that the number of iterations is constant. Through empirical observations, $T=500$ is sufficient to achieve convergence in \name{} for all AdNN models. Note that sensitivity cannot be compared using the magnitude of $c$ as only BranchyNet and RANet use normalization filters, whereas BlockDrop does not, which makes the optimal $c$ of BranchyNet and RANet larger (see Table \ref{tab:HP}). 
To measure the magnitude of perturbation for a set of c, T we measure the average squared difference between the testing input and the input image, which is defined as follows:
$$ \textrm{Average squared difference} = \frac{1}{N} \sum_{i=1}^{N} (x_i - f_i)^2  $$
where $x_i$ is the input image, and $f_i$ is its corresponding testing input. 
Figures \ref{fig:robustness_cifar10} and \ref{fig:robustness_cifar100} show the average percentage increase in energy consumption versus the average squared difference on the CIFAR-10 and CIFAR-100 datasets. We observe that for both datasets compared to BlockDrop, BranchyNet and RANet are more sensitive to the perturbation magnitude. BlockDrop’s lower sensitivity is mainly due to BlockDrop’s Policy Network, which provides more refined control over energy consumption; hence BlockDrop is more robust than the other two AdNNs. Additionally, we can see the potency of the Estimator model for every AdNN model. A direct proportionality between the average increase in energy consumption and the average squared difference is observed in all the AdNN models, which is evidence that the Estimator model is successful in imitating the energy consumption of each AdNN. Additionally, we observe that \name{} performs very similarly for both CIFAR-10 and CIFAR-100 datasets for BlockDrop and BranchyNet. Whereas for RANet for CIFAR-10, the energy spike induced is much higher than that for CIFAR-100, indicating that the CIFAR-100 RANet is more robust than the CIFAR-10 version.

\begin{figure}[!t]
    \centering
\begin{subfigure}[t]{0.23\textwidth}
        \includegraphics[width=\textwidth]{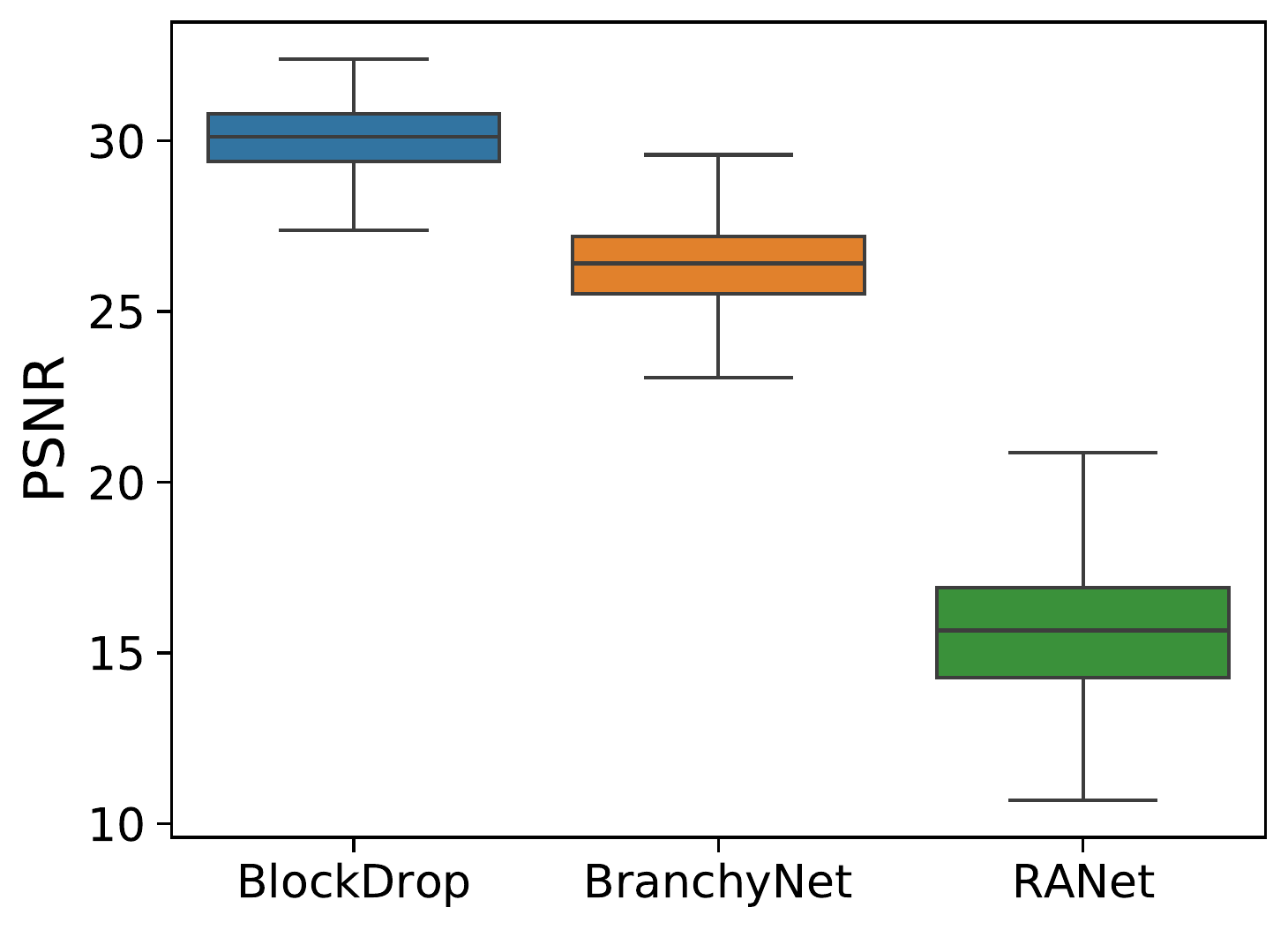}
        \caption{PSNR between original and generated images }
         \label{fig:PSNR_RQ3}
    \end{subfigure}
    \begin{subfigure}[t]{0.23\textwidth}
        \includegraphics[width=\textwidth]{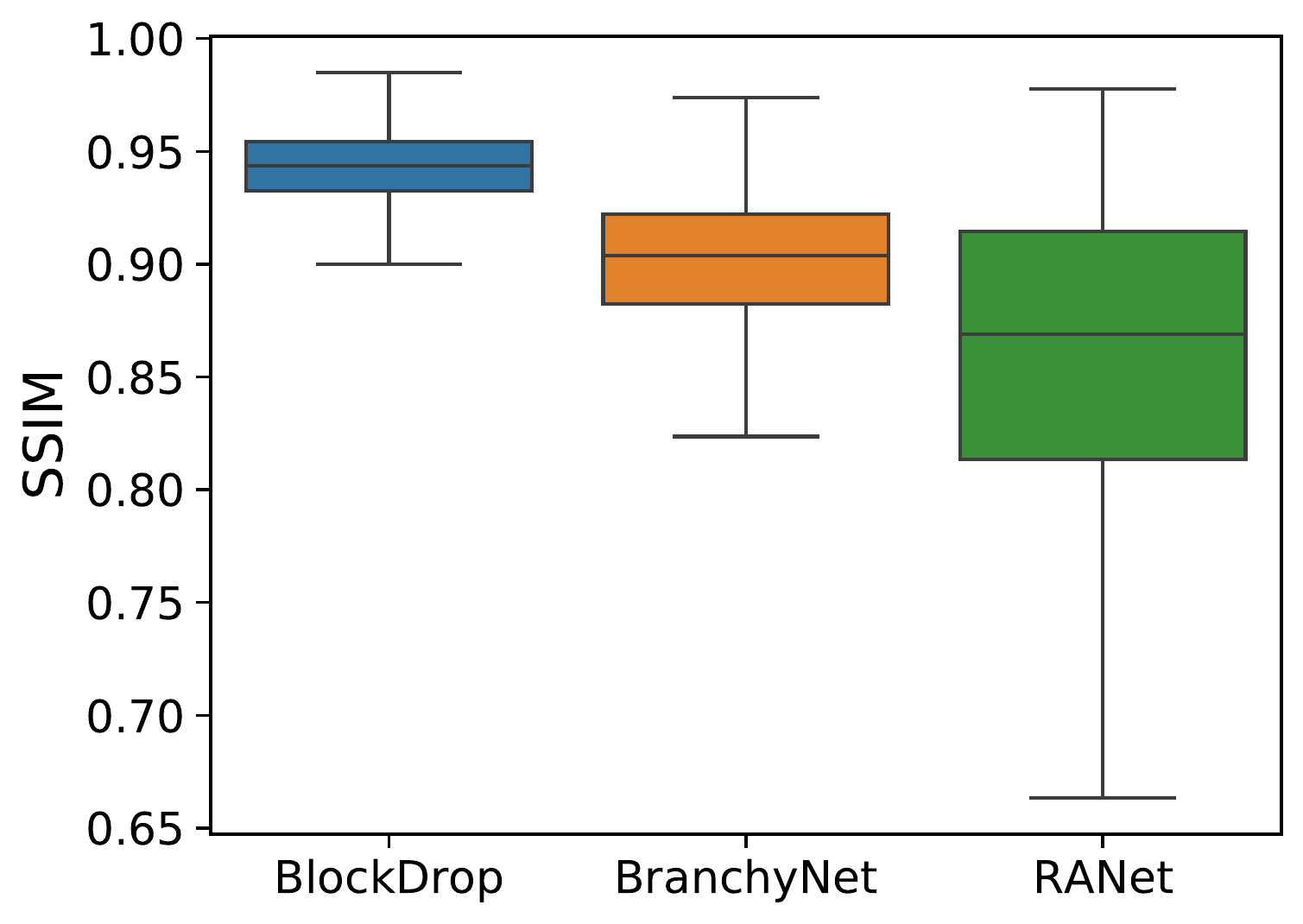}
        \caption{SSIM between between original and generated images.}
        \label{fig:quality2}
    \end{subfigure}
     \caption{Quality of the generated images for each AdNN model for CIFAR-10 dataset}
    \label{fig:quality_new}
\end{figure}

\begin{figure}[!t]
    \centering
\begin{subfigure}[b]{0.23\textwidth}
        \includegraphics[width=\textwidth]{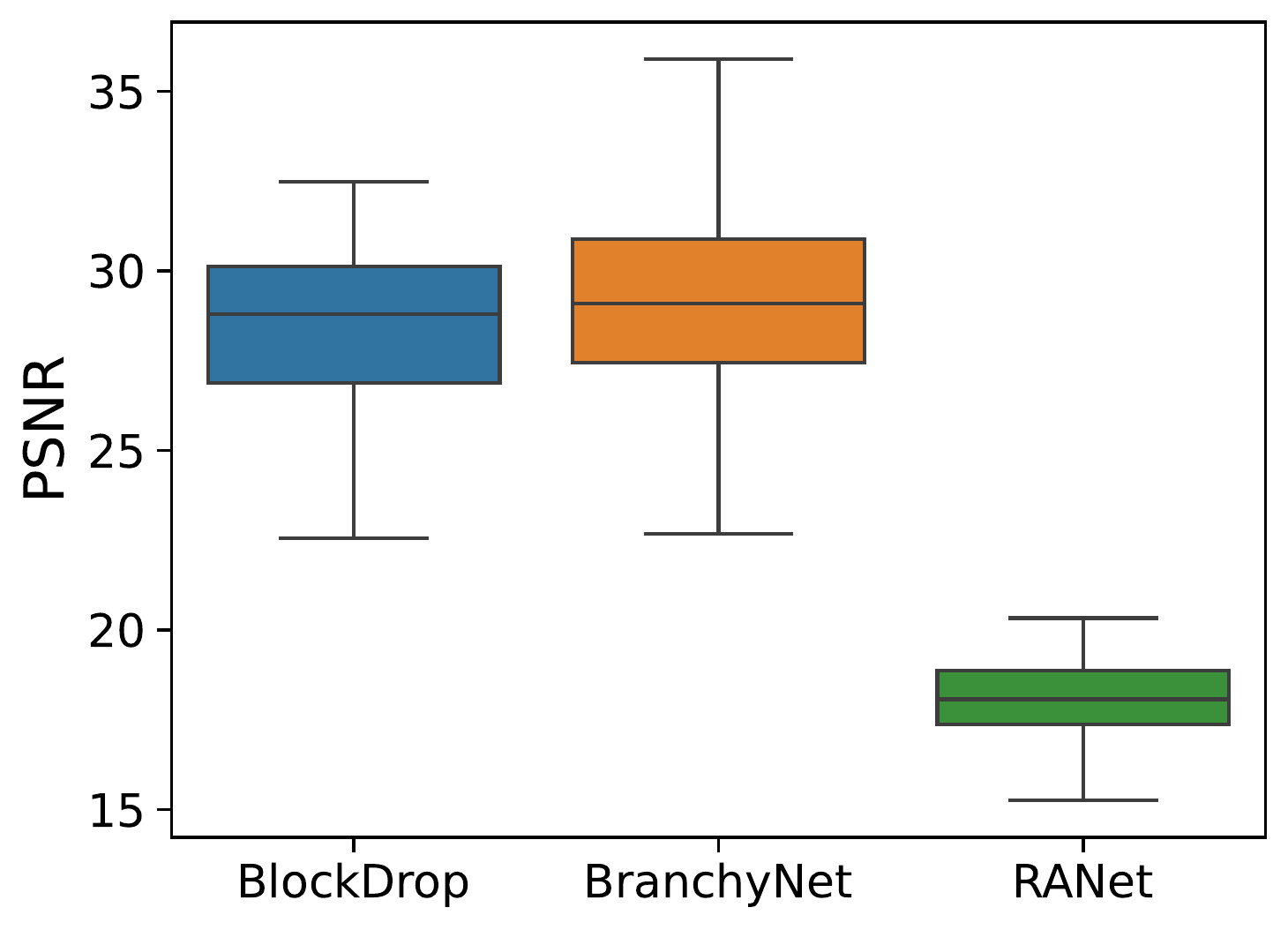}
        \caption{PSNR between original and generated images }
         \label{fig:PSNR_RQ3}
    \end{subfigure}
    \begin{subfigure}[b]{0.23\textwidth}
        \includegraphics[width=\textwidth]{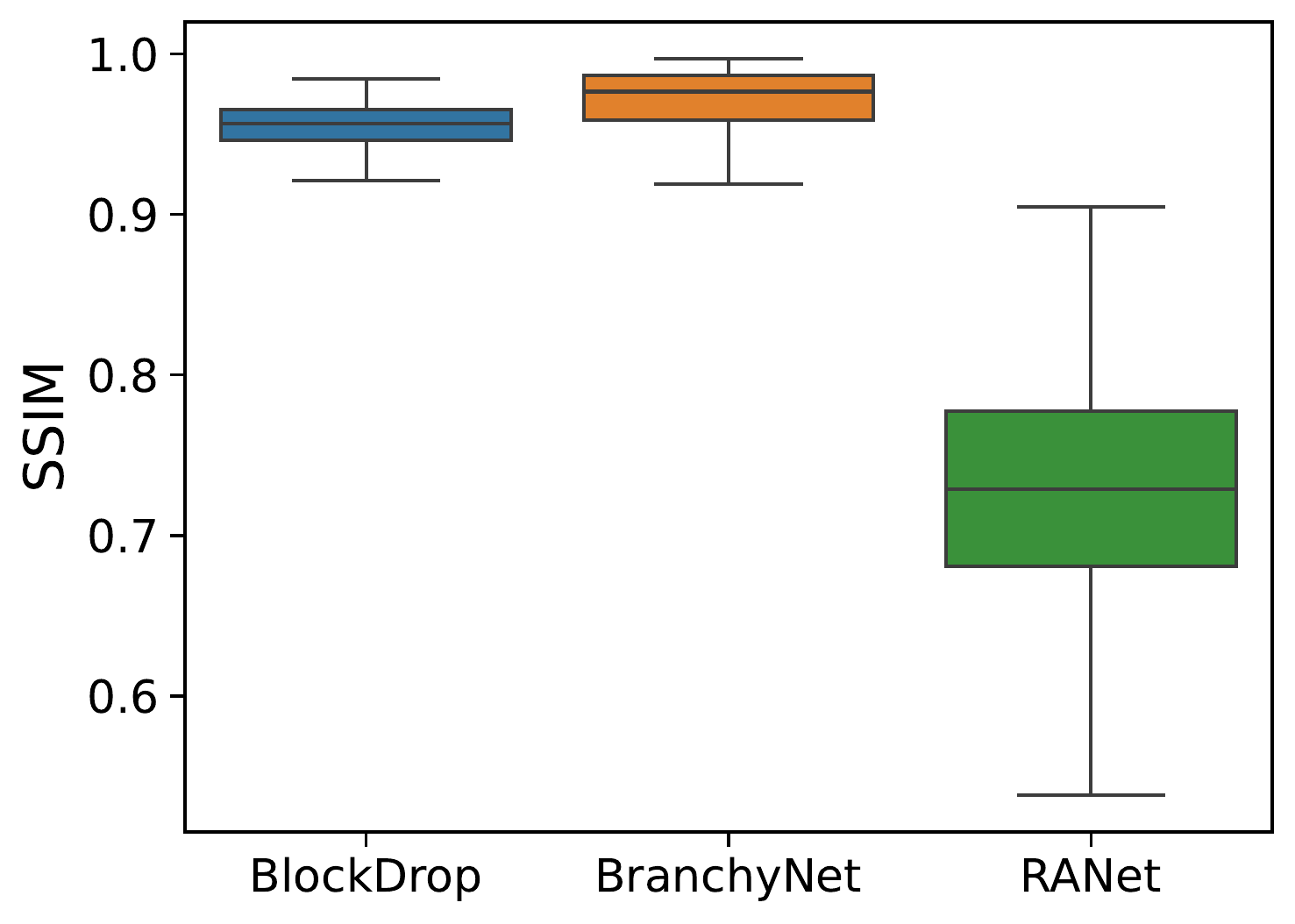}
        \caption{SSIM between between original and generated images.}
        \label{fig:quality2}
    \end{subfigure}
     \caption{Quality of the generated images for each AdNN model for CIFAR-100 dataset}
    \label{fig:quality_new_c100}
\end{figure}


\subsubsection{RQ3: Quality}
\label{sec:quality}


	

In this section, we evaluate the quality of the perturbation generated by Input-based testing of \name{} with the hyper-parameters set to the values given in Section \ref{subsec:setup} using Peak Signal to Noise ratio (PSNR) \cite{psnr} and Structural Similarity Index (SSIM) \cite{SSIM}. Both of these metrics are used in the industry to measure the image quality of noisy images. SNR of an image can be represented by,$$ SNR = \frac{\mu_{image}}{\sigma_{image}}$$ where $\mu_{image}$ is the mean value of the image pixels and $\sigma_{image}$ is the error value of the pixel values. For PSNR, the highest value of the image pixels is used instead of the mean value.
The Structural Similarity Index (SSIM) is a perceptual metric that quantifies the image quality degradation caused by processing such as data compression or by losses in data transmission.
Higher values for SSIM and PSNR indicate higher quality test inputs. 

Figure \ref{fig:quality_new} and Figure \ref{fig:quality_new_c100} show the values of SSIM and PSNR between the testing inputs and the original images for CIFAR-10 and CIFAR-100 datasets. For both datasets, we see that SSIM for the generated testing inputs is similar for all the target AdNNs. Whereas PSNR for RANet is worse but still comparable to BlockDrop and BranchyNet. We conclude that most of the inputs generated through Input-based testing are of high quality; even if some test inputs might have noise, they are structurally similar to the original inputs.

\npdecimalsign{.}
\nprounddigits{2}

\begin{table}[h]
\centering
\scriptsize
\setlength{\tabcolsep}{0.3em}
{\renewcommand{\arraystretch}{1}
\caption{Effect of corruptions on \name{}. Average percentage increase in energy consumption for all AdNN models for various corrupted inputs from CIFAR-10 and CIFAR-100.}
    \label{tab:Robustness}
\begin{tabular}{|c|c|n{5}{2}|n{5}{2}|n{5}{2}|n{5}{2}|}
\hline
AdNN & Dataset & Normal & Frost & Fog & Snow \\
\hline
\multirow{2}{*}{RANet } 
  & CIFAR-10 & 929.756286891818 & 2085.364467238083 & 2027.1972381639227 & 2099.40819573178\\
\cline{2-6} & CIFAR-100 & 340.75359729076314 & 6.9775027539506365 & 13.542675484632257 & 25.51062916892334 \\
\hline
\multirow{2}{*}{BranchyNet} 
 & CIFAR-10 & 283.05016382733407 & 475.0805230629547 & 355.69764238270096 & 323.71718253094576\\
\cline{2-6} & CIFAR-100 & 278.17356659741966 & 197.57622281220145 & 374.6236015924043 & 360.3738774504222\\
\hline
\multirow{2}{*}{BlockDrop } 
  & CIFAR-10 & 72.77580615655118 & 108.37144434633245 & 89.60675827642109 & 111.8001840439201 \\
\cline{2-6} & CIFAR-100 & 17.49923808054768 & 24.909844994672394 & 21.49132322669614 & 36.949479104537964 \\
\hline
\end{tabular}}

\end{table}
\subsubsection{RQ4: Robustness}
\name estimates energy consumption based on the training data, which can be impacted by distribution shift~\cite{quionero2009dataset}. Therefore, to evaluate the robustness of \name against distribution shifts, we have analyzed the behavior of \name{} against Practical corruptions. 
Practical corruptions (\textit{e.g.,} fog, snow etc.) are frequently noticed, mainly when we use mobile phones or autonomous vehicles to take images.
For this purpose, we have used real-world common corruption techniques \cite{hendrycks2019robustness}.  

Table \ref{tab:Robustness} shows the average percentage increase in energy consumption for the testing inputs generated using the original and corrupted images of CIFAR-10 and CIFAR-100 by \name{} in Input-based testing. We picked the corruption classes of fog, frost, and snow due to their natural occurrence. In general, the corrupted images do not hinder the performance of \name{} except for the CIFAR-100 version of RANet. In other cases, the increase in energy consumption achieved by \name{} using corrupted images is, in fact, higher than that achieved using the original CIFAR-10 and CIFAR-100 datasets. This is mainly due to the common corruptions  \cite{hendrycks2019robustness}, introducing better initialization spots (white areas). For CIFAR-100 RANet, \name{} manages to increase the energy consumption slightly. However, similar to the Universal testing case, inputs with high noise are classified as road or shrew with high confidence, which leads to lower performance in comparison to other settings.

Additionally, through these results, we further notice the high stability of the Estimator model in approximating the shortcomings of the target AdNNs.  \name{} can generate high energy-consuming testing inputs despite the corruption of images. 
While there is some variance in how \name{} behaves when provided with an image with different corruption classes; the median energy consumption increment is consistent for all target AdNNs.

%% file: discussion.tex
\section{Increasing Energy Robustness}

In this section, we demonstrate two ways to increase the energy robustness of AdNNs. In both ways, we use prior \name{} generated inputs to detect new \name{} generated energy-surging inputs for AdNNs. For detecting Universal test inputs, we use input filtering method based on pixel values, while for detecting input-based test inputs, we use gradient-based input detection.


\noindent \textbf{Input Filtering.}
As we can notice that noisy samples generated from Universal testing can increase energy consumption by a significant amount, hence it is essential to adapt the AdNN mechanism against these noisy images. For traditional DNNs, highly noisy images consume the same energy as the standard images and are no threat to the robustness of the DNN (the object is not visible in those noisy images). To adapt AdNNs against high energy-consuming images, we propose to include a filter in AdNNs.

As a filter, we have created a ResNet model (binary classifier) with only six residual blocks. The classifier can classify into two categories: \textbf{normal input} and \textbf{energy-consuming noisy input}. For each AdNN, We have trained the ResNet model with 2500 normal dataset images and 2500 high energy-consuming noisy images, creating three different trained ResNet models. For testing the models, 1000 images have been used (500 from each class). Our results show that the filter can identify each test image with the correct class (100\% accuracy) for all three AdNNs for both datasets. The results confirm that we can filter out high energy-consuming noisy images with a model whose energy consumption is low.

\noindent \textbf{Gradient-based Input Detection.}
In contrast to detecting adversarial inputs similar to the samples generated by the Universal testing mode, adversarial inputs similar to the samples generated by the  Input-based testing mode are harder to differentiate from benign inputs. Additionally, the energy constraint on such a detection system is a significant challenge. Therefore, the detection technique must consume significantly low energy with respect to the energy consumed by Input-based testing inputs during inference. To address these challenges, we propose a gradient-based adversarial input detection mechanism that uses partial inference from the AdNN. 

In this detection mechanism, we leverage the behavior of the energy-saving mechanism within various AdNNs. These mechanisms, in general, try to ensure that the difference between an intermediate output and a predefined condition is large, which will deactivate certain parts of the AdNN. In other words, if the loss function of intermediate outputs is large for any input, the energy consumption will be low \cite{MirazILFO}. Therefore, if the gradients of the intermediate loss function with respect to the weights are large, the input is more likely to be a benign input. Hence, we only need partial inference (weight gradients of an initial layer) from the AdNNs, and a linear SVM \cite{cortes1995support} to detect the energy-surging inputs, where both are low energy consuming steps. So, the energy impact of our detection component is significantly low. If the input is predicted as energy-surging by the detector, the inference is stopped early.

To evaluate our detection component, we generate input-based testing inputs for CIFAR-10 and CIFAR-100 training and test datasets for all three AdNNs. Next, we calculate the gradients of the weights with respect to the intermediate loss function for the training section of datasets. For all three AdNNs, we consider the weights of the first layer of the AdNN. Specifically for BlockDrop, we calculate the intermediate loss function of the policy network, whereas, for RANet and BranchyNet, we use the first exit's loss function (Section \ref{subsec:setup}). After calculating the weight gradients for the training section of datasets, we label them as either original or testing inputs and use them for training an SVM binary classification model for each AdNN. Finally, we test the SVM classifiers on the gradients generated by the original and input-based testing input for the testing section of datasets.


\begin{table}[h]
\centering
\scriptsize
\setlength{\tabcolsep}{0.3em}
{\renewcommand{\arraystretch}{1}
\caption{Detection Accuracy (\%), AUC score, Accuracy Drop percentage, Adversarial Energy Decrease Percentage, and Benign Energy Increase Percentage of the Gradient-based input detection incorporated in to various AdNN models}
    \label{tab:Def}
\begin{tabular}{|*{56}{c|}}
\hline
AdNN & Dataset & Detection (\%) & AUC & Acc Drop(\%) & Adv Eng Dec (\%) & Ben Eng Inc (\%) \\
\hline
\multirow{2}{*}{RANet }  
  & CIFAR-10 & 92.50 & 0.924&11.50 &77.10 &26.80\\
\cline{2-7}& CIFAR-100 & 81.60& 0.816 &3.90 & 84.30&17.74\\
\hline
\multirow{2}{*}{BranchyNet} 
 & CIFAR-10 & 99.90&0.999 & 0.01& 84.40&17.90\\
\cline{2-7}& CIFAR-100 & 99.80 & 0.998& 0.01&87.50 &15.60\\
\hline
\multirow{2}{*}{BlockDrop } 
  & CIFAR-10 & 94.39&0.943& 3.30 &95.00&6.25 \\
\cline{2-7}& CIFAR-100 & 66.90 & 0.666&30.40 &92.80&7.60 \\
\hline
\end{tabular}}

\end{table}

Table \ref{tab:Def} shows the detection accuracy (\%) and AUC score \cite{10.1016/S0031-3203(96)00142-2} of our gradient-based input detection technique. AUC score computes the area under the Receiver Operating Characteristic Curve (ROC AUC) \cite{10.1016/S0031-3203(96)00142-2} and measures the efficacy of any binary classifier, with higher AUC scores corresponding to better classifiers. The results show that in 5 out of 6 scenarios, both AUC score is higher than 0.8 and the detection accuracy is higher than 80 percent, showing our approach's efficacy. Additionally, we also report the accuracy drop of the AdNNs due to false positives from the detection system. We observe that the accuracy drop is minimal in all cases except for the CIFAR-100 BlockDrop model.

Furthermore, to demonstrate the benefits of our gradient-based detection system from the energy perspective, we also report the average energy decrease percentage for an adversarial input (Adv Eng Dec) and average energy increase percentage for a benign input (Ben Eng Inc). We observe that our detection system can significantly reduce energy consumption induced by adversarial inputs (up to 95\%) while introducing minimal energy burden as evidenced by a low increase in consumption for benign inputs.

\section{Discussion}
\label{sec:discussion}
We discuss the alternative defense for AdNNs, the adaptability of AdNNs on different datasets, and the relationship between block activation and energy consumption of AdNNs. Also, we discuss correlation between measured and estimated energy consumption, and the correlation between energy consumption increased by different techniques.

\subsection{Alternative Defense.}
We have investigated the application of adversarial training as an alternative defense mechanism against \name{} generated energy-surging inputs.
To understand the effect of adversarial training on AdNNs, we use the \name{} generated testing inputs for CIFAR-10 dataset to retrain the original AdNNs. 
We used Input-based testing inputs generated from 1000 images of the CIFAR-10 training dataset as the training set and retrained the BranchyNet and RANet AdNNs for 150 epochs with a learning rate same parameters as the initial training. 
We generate the test set using a batch of 600 images from the CIFAR-10 test dataset. 
We found that adversarial training does not increase the energy robustness for all AdNN models. Specifically, RANet is easier to improve using adversarial training compared to BranchyNet. Due to space constraints, we have reported the results for Adversarial training in our website \footnote{https://sites.google.com/view/ereba/home}.

\begin{figure}[t]
    \centering
    \begin{subfigure}[t]{0.23\textwidth}
        \includegraphics[width=\textwidth]{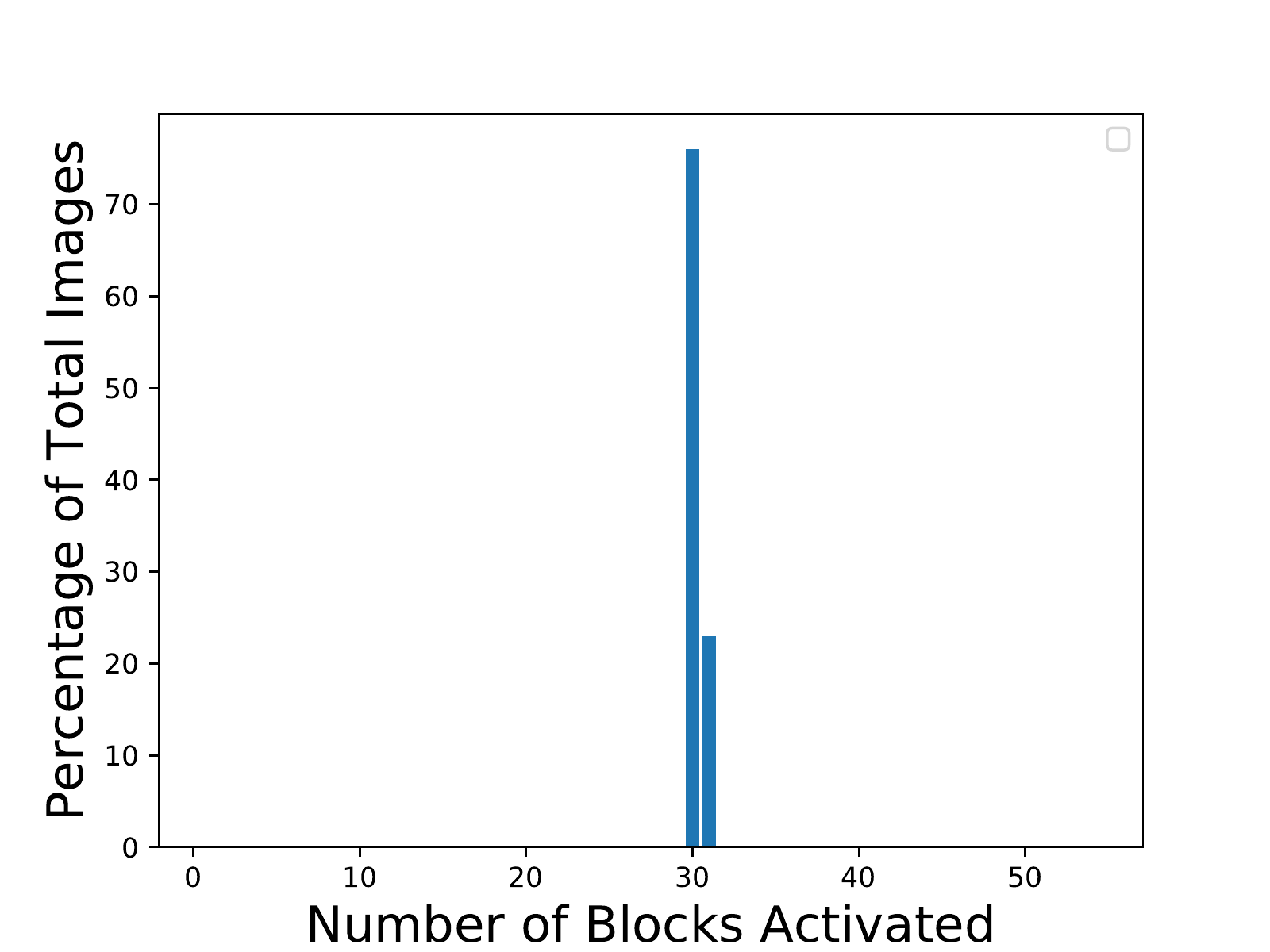}
        \caption{Adaptability of BlockDrop (RN-110) on CIFAR-100 Dataset }
        \label{fig:adaptabilityBDC100}
    \end{subfigure}
    \begin{subfigure}[t]{0.23\textwidth}
        \includegraphics[width=\textwidth]{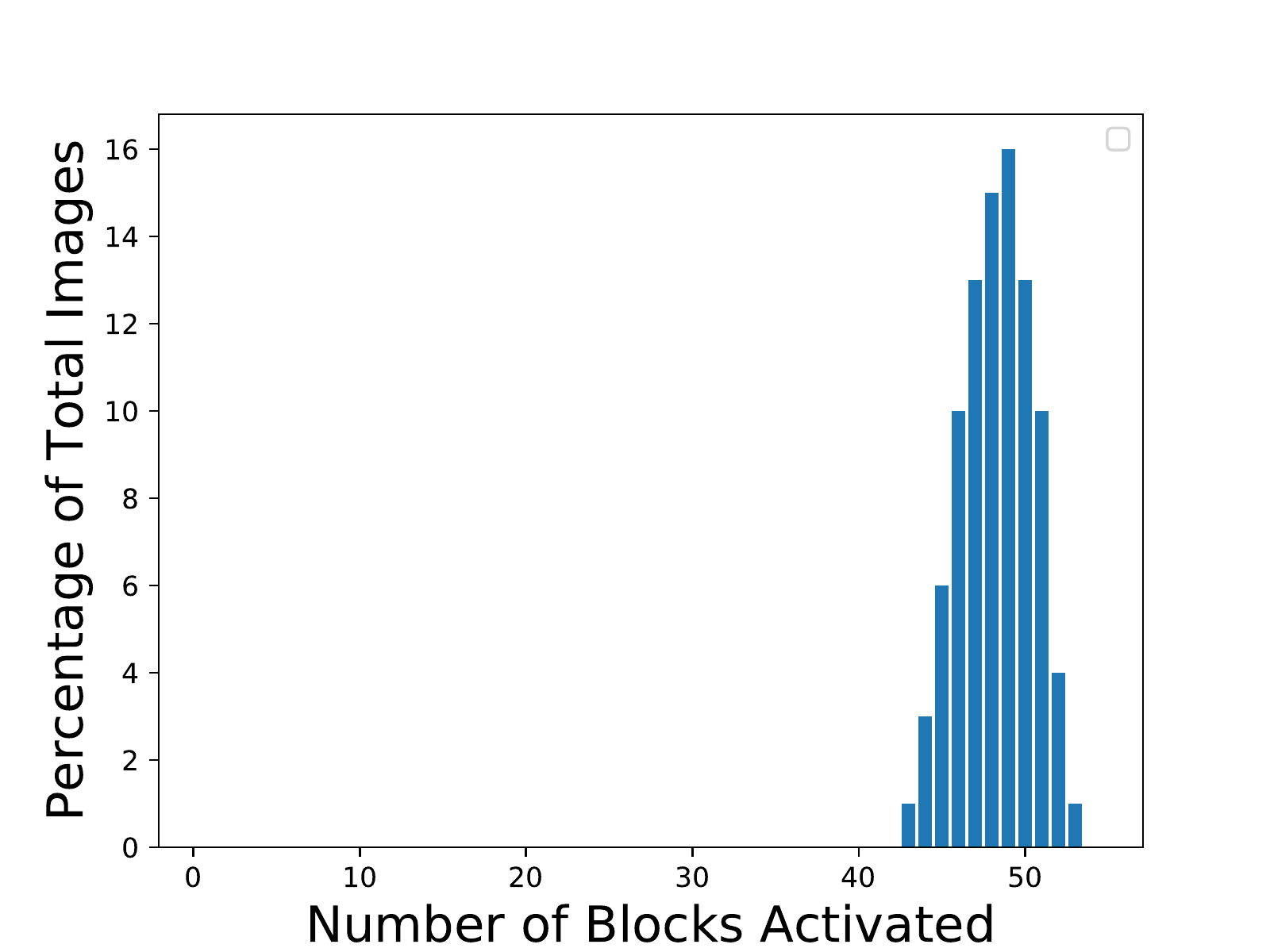}
        \caption{Adaptability of SkipNet on CIFAR-100 Dataset }
        \label{fig:adaptabilitySNC100}
    \end{subfigure}
    \begin{subfigure}[t]{0.23\textwidth}
        \includegraphics[width=\textwidth]{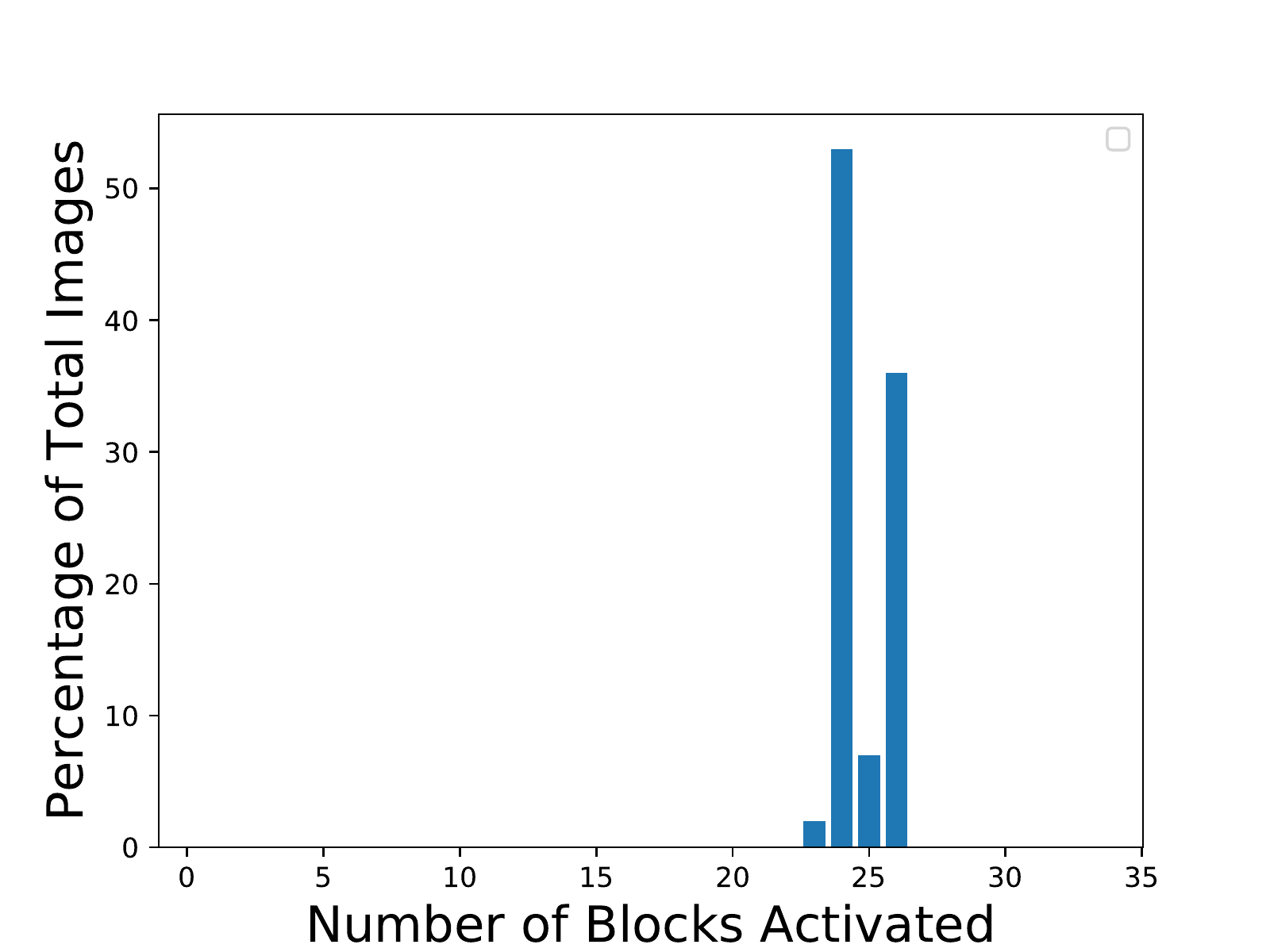}
        \caption{Adaptability of BlockDrop (RN-110) on ImageNet Dataset }
        \label{fig:adaptabilityInet}
    
    \end{subfigure}
    \begin{subfigure}[t]{0.23\textwidth}
        \includegraphics[width=\textwidth]{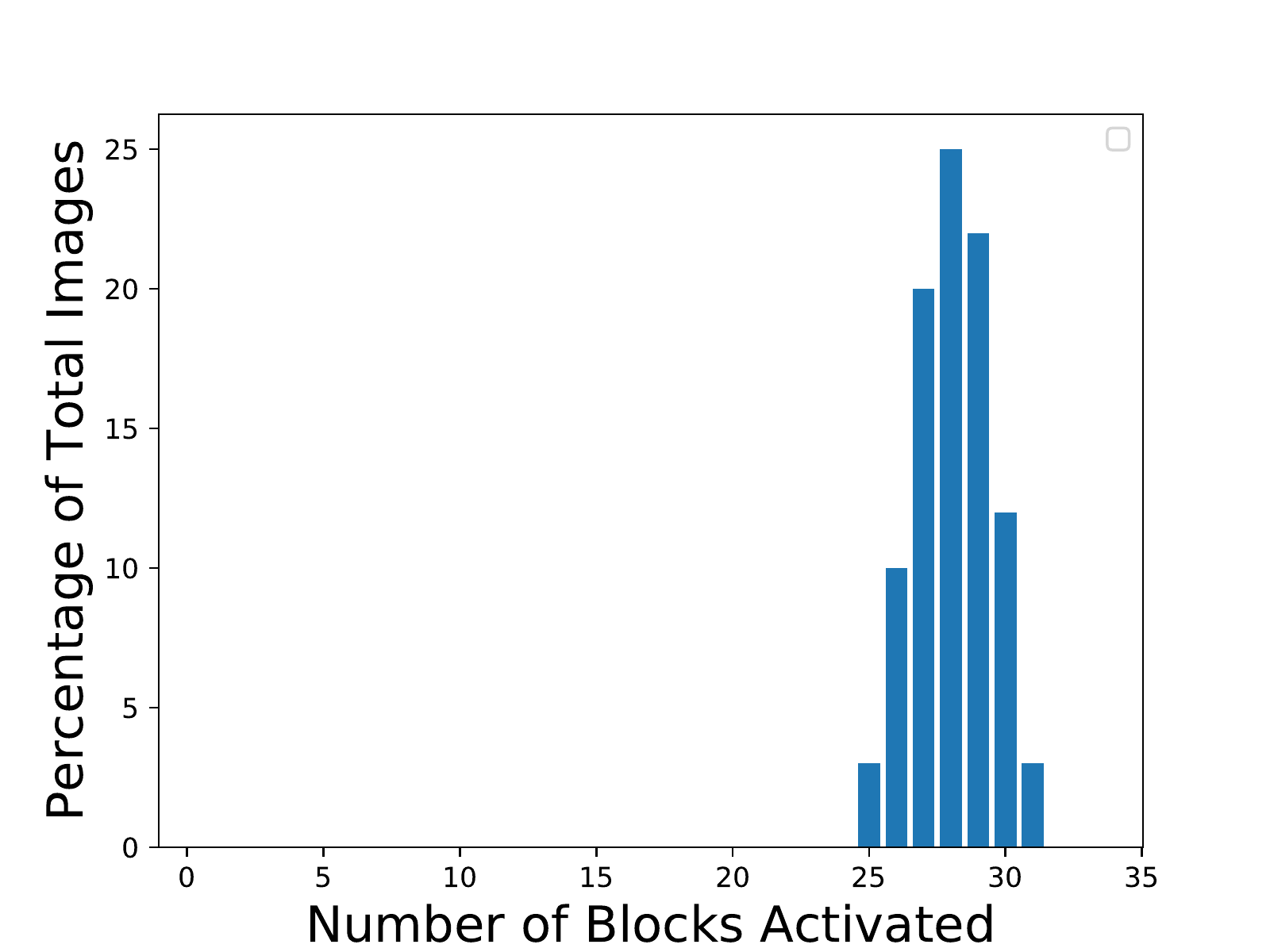}
        \caption{Adaptability of SkipNet on ImageNet Dataset }
        \label{fig:adaptabilitySNInet}
    \end{subfigure}
    \caption{Adaptability of AdNNs on ImageNet and CIFAR-100 dataset.}
\end{figure}

\begin{figure*}[t]
    \centering
    \begin{subfigure}[b]{0.26\textwidth}
        \includegraphics[width=\textwidth]{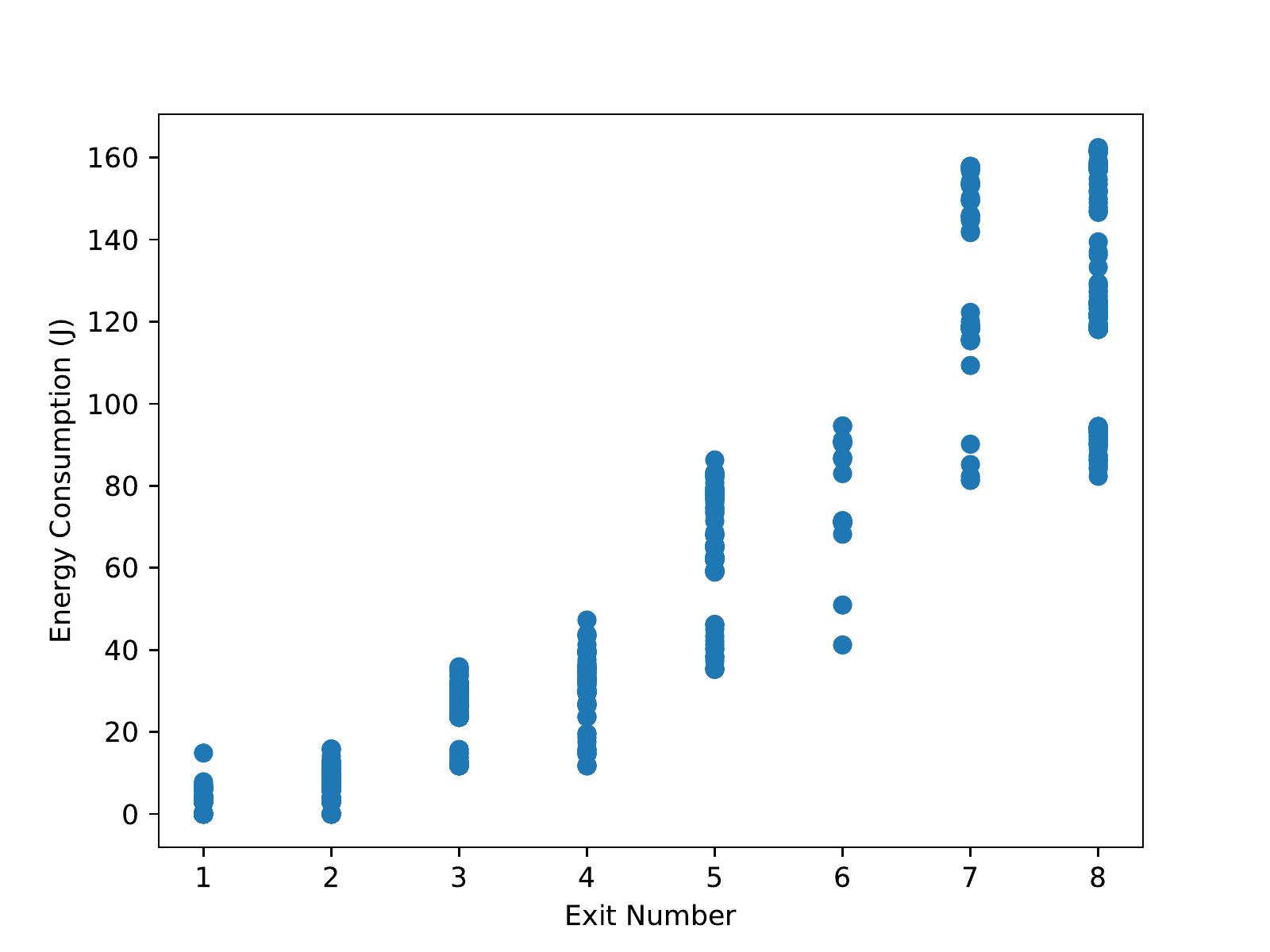}
        \caption{}
    \end{subfigure}
    \begin{subfigure}[b]{0.26\textwidth}
        \includegraphics[width=\textwidth]{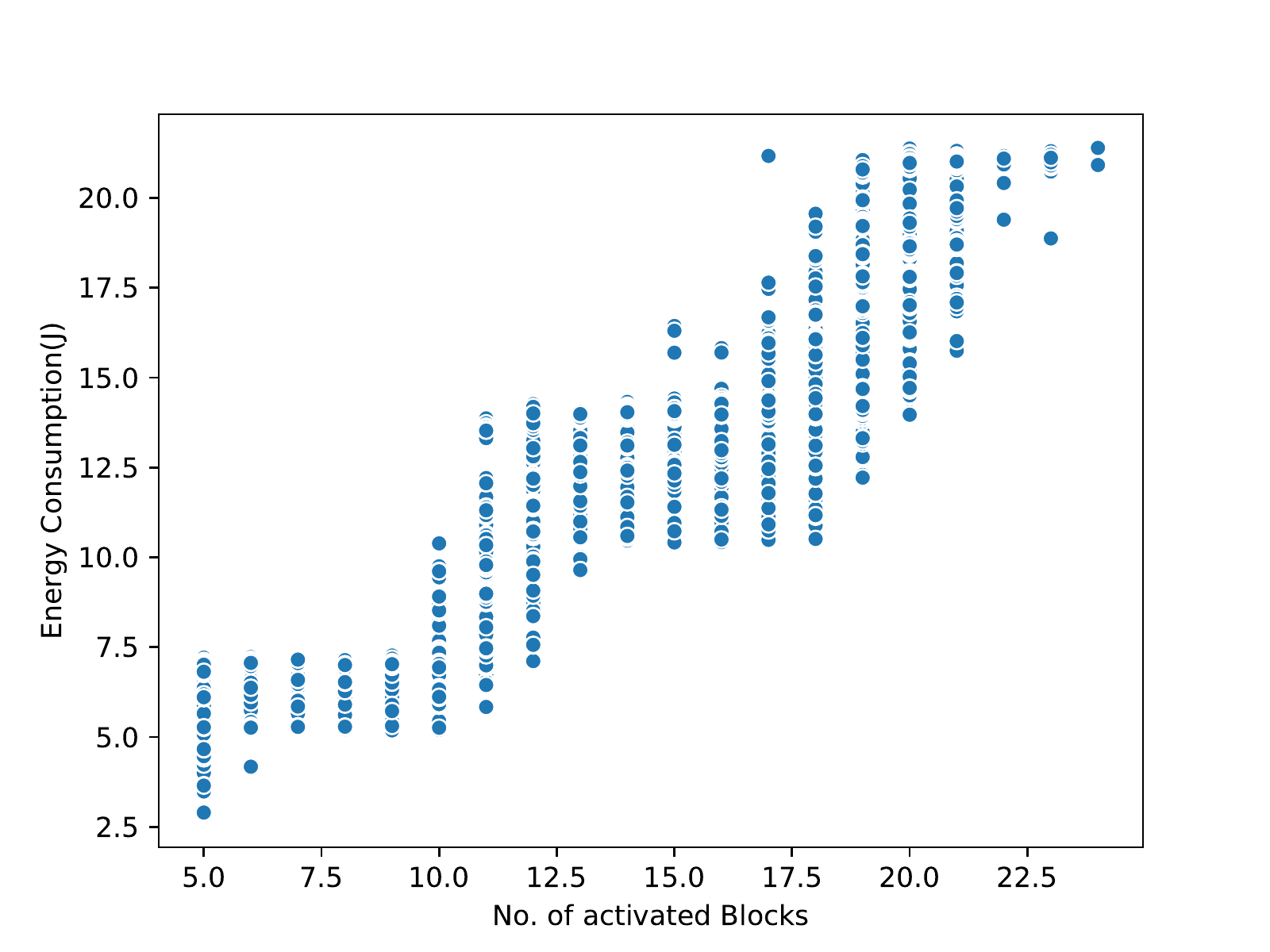}
        \caption{}
    \end{subfigure}
    \begin{subfigure}[b]{0.26\textwidth}
        \includegraphics[width=\textwidth]{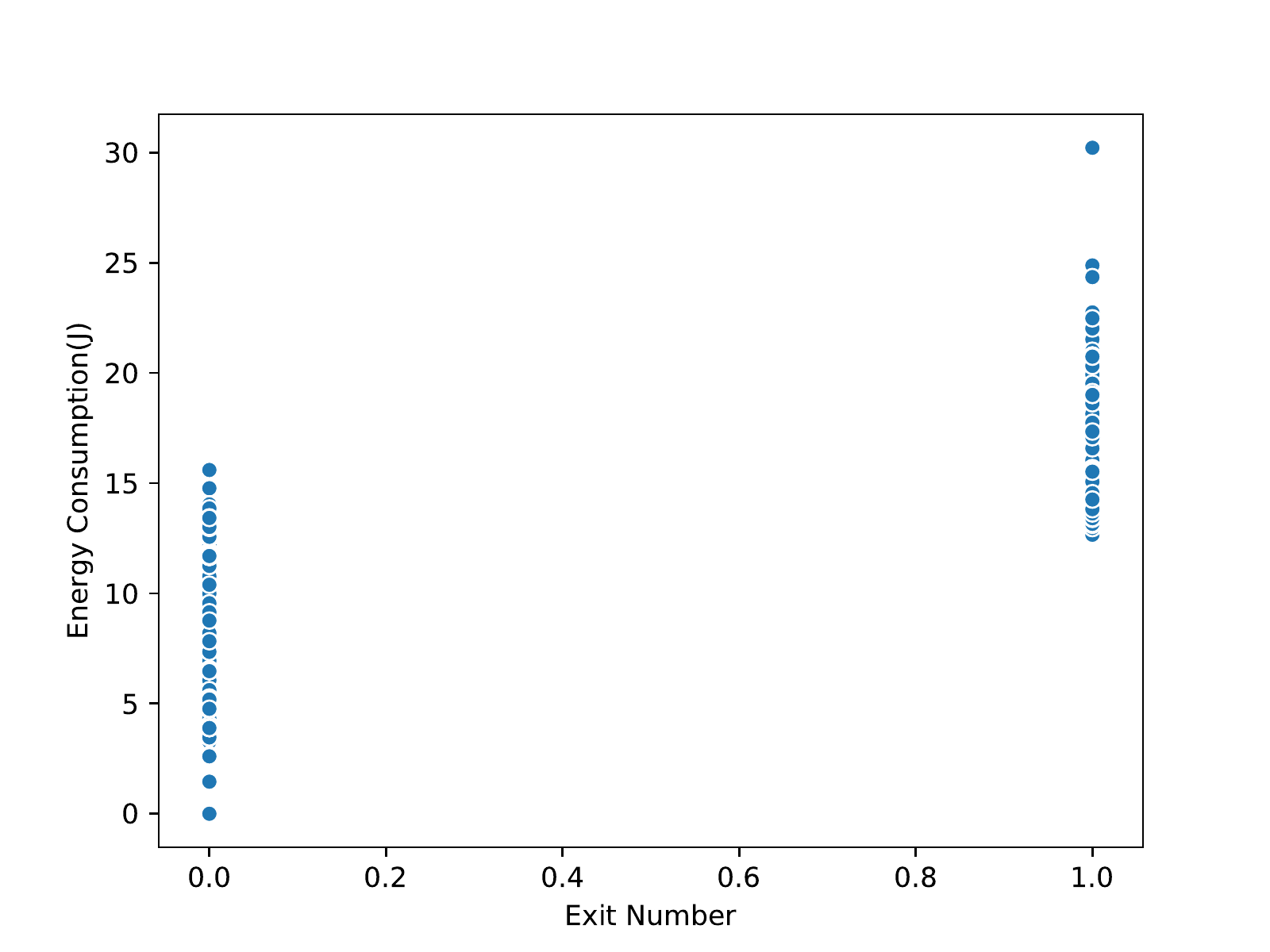}
        \caption{}
    \end{subfigure}
    \caption{Energy consumption of (a) RANet, (b) BlockDrop,  (c) BranchyNet for CIFAR-10 Training Set}
    \label{fig:dicussion}
\end{figure*}

\subsection{Adaptability of AdNNs}
\label{sec:adaptability}
In our observations, we see that AdNNs may not be adaptive under all circumstances. Each AdNN can decrease its FLOPs count; however, this may not always result in concrete energy consumption patterns. Figures \ref{fig:adaptabilityBDC100}, \ref{fig:adaptabilitySNC100}, \ref{fig:adaptabilityInet}, and \ref{fig:adaptabilitySNInet} show BlockDrop (ResNet-110) and SkipNet models' adaptability on CIFAR-100 and ImageNet datasets. We can see that the BlockDrop's adaptability (the difference between the highest and lowest number of activated blocks) for both datasets is limited (less than 4). For SkipNet, the range of adaptability is better than BlockDrop's range.

\subsection{Comparison to Misclassification Attacks}
\label{sec:pattern_adnn}
We can assume that the energy-saving mechanisms of Conditional-Skipping networks like BlockDrop would classify an image into $N - 1$ categories. $N$ is the number of blocks in the DNN, with each image in category $i$ activating $i+1$ (the first block is always active) blocks. Whereas for the case of Early-Termination networks such as BranchyNet and RANet, they classify an image into $E$ categories where $E$ is the number of exits in the network. However, due to noise from the system environment, our Estimator model cannot differentiate between all the classes. Figure \ref{fig:dicussion} shows the scatter plots between these image classes and their energy consumption for each AdNN model. We can see the step-wise pattern in the plot for all the AdNN models. The Estimator model is trained to learn this pattern of energy consumption of AdNNs.

\subsection{Correlation between Actual Energy Consumption and Estimated Energy Consumption}

To illustrate the correlation between energy consumption predicted by the estimator model and original energy consumption, we use the Pearson Correlation Coefficient (r)~\cite{benesty2009pearson} and correlation p-value. If two sets of values are correlated, the r value would be significant, and the p-value will be low.

For CIFAR-100 dataset, the r values are 0.38, 0.17, and 0.31 for  RANet, BlockDrop, and BranchtNet models, respectively, where all the p-values are less than 0.0005. These results conclude that the values are correlated. 

For CIFAR-10 dataset, the value of r for RANet is 0.021; however p-value is 0.03; therefore, it is more likely that the values are correlated. For BlockDrop model, the value of r and p-value are 0.004 and 0.66, which suggests that the values are less likely to be correlated. But if we consider only the inputs whose energy consumption is higher than the 75th percentile value, the p-value becomes 0.14, suggesting a correlation. Therefore, if the estimator model can accurately predict high energy-consuming inputs (i.e., differentiate clearly between low/mid and high energy-consuming inputs), we can use the estimator model to generate energy-expensive testing inputs.

\subsection{Correlation of Increase of Energy Consumption between Different Techniques}

In this section, we try to explore the correlation between energy consumption modified by Input-based testing and baseline techniques.
We use Pearson Correlation for that purpose. Pearson Correlation is one of the metrics that can find the strength of the relationship between two variables. For CIFAR-100 data, we show the Table \ref{tab:correl_eng} that represents the Pearson Correlation Coeff (r) and p-value between the percentage of energy consumption increased by Input-based testing inputs and energy consumption increased by baseline technique generated inputs. It can be noticed that for most of the cases, the energy increase percentages are less likely to be correlated. Only for BranchyNet, we can find a significant negative correlation between Input-based and Perturbation-induced energy consumption increase.

\begin{table}[h]
\centering
\scriptsize

\setlength{\tabcolsep}{0.3em}
{\renewcommand{\arraystretch}{1}
\caption{Correlation between the percentage of energy consumption increased by Input-based testing inputs and percentage energy consumption increased by baseline technique generated inputs for different models for CIFAR-100 data}
    \label{tab:correl_eng}
 \begin{tabular}{|l| r| r| r|r|} 
 \hline
 \diagbox[width=15em]{\rule{0mm}{3mm}Models}{\rule{0mm}{3mm}Baseline} & r(Perturb) & p-value(Perturb) & r(SURRG) & p-value(SURRG)  \\  [0.1ex]
 \hline
 RANet & 0.142 &  0.328& -0.04& 0.976 \\
 \hline
 BlockDrop & -0.007 & 0.908 & -0.037&0.544 \\
 \hline
 BranchyNet & -0.296 & 0.133 & -0.118&0.556 \\
 \hline

\end{tabular}}


\end{table}

%% file: related.tex
\section{Related Works}
\label{sec:related_work}
\textbf{AdNNs.} 
 Among Conditional-skipping models, 
Hua \textit{et al.} \cite{hua2019channel} and Gao \textit{et al.} \cite{gao2018dynamic} explore channel gating to determine computational blind spots for channel-specific regions unessential to classification. Liu \textit{et al.} \cite{liu2018dynamic} propose a new type of AdNN which utilizes reinforcement learning to achieve selective execution of neurons. SkipNet \cite{wang2018skipnet} uses gating techniques to skip residual blocks.
On the other hand, Graves \textit{et al.} \cite{graves2016adaptive}, Figurnov \textit{et al.} \cite{figurnov2017spatially}, and Teerapittayanon \textit{et al.} \cite{teerapittayanon2016branchynet} propose SACT and BranchyNet respectively, which are Early-termination AdNNs. SACT terminates the computation within a residual block early based on intermediate outputs, while BranchyNet uses separate exits within network for early termination.
Cascading multiple DNNs with various computational costs through a single computation unit to decide which DNN to execute has been proposed. The cascading models use various techniques such as termination policy \cite{bolukbasi2017adaptive}, reinforcement learning \cite{guan2017energy}, and gating techniques \cite{nan2017adaptive} to achieve early termination.

\textbf{Adversarial Examples.} Adversarial Examples are the synthesized inputs that is able to modify the prediction of the ML model. Szegedy \textit{et al.} \cite{szegedy2013intriguing} and Goodfellow \textit{et al.} \cite{goodfellow2014explaining} propose white-box adversarial attacks on convolutional neural networks.
  Papernot \textit{et al.} \cite{papernot2017practical} have used surrogate model to attack a DNN in black-box setting. Liu \textit{et al.} \cite{liu2016delving} use ensemble of multiple white-box models to generate adversarial examples, which can attack black-box models. Ilyas \textit{et al.} \cite{ilyas2018black} use evolutionary search strategies to estimate the gradient of a model to attack black-box models.

 However, all these attacks focus on changing the prediction and do not concentrate on increasing test time. ILFO \cite{MirazILFO} is the first work to attack a DNN by increasing the energy consumption of the model. However, ILFO uses white-box setting and does not have transferability. Therefore, ILFO can not be used for black-box attack.\balance

 Next, DeepSloth~\cite{hong2020panda} uses modified PGD to attack against Early-termination AdNNs using the confidence scores in each exit. However, DeepSloth can not be used against Conditional-skipping AdNNs. Also, DeepSloth provides a study about the transferability of the attack. The study considers the efficacy of the Early-termination models as the transferability metric. However, we propose a more systematic transferability study by introducing of metrics like ETP and ITP.


\textbf{DNN Testing.}
Multiple testing methods have been proposed recently to test DNNs. DeepGauge \cite{ma2018deepgauge} is proposed based on a  test criteria set that
verifies the corner neuron activation values. DeepXplore \cite{pei2017deepxplore} proposes to cover each neuron's binary activation status and use neuron coverage to test DNNs.
 DeepTest \cite{tian2018deeptest} tests autonomous driving cars by using neuron coverage. Recently, DeepHunter \cite{xie2019deephunter} proposes to use coverage-guided fuzz testing on DNNs. \name evaluates the energy-robustness of AdNNs in a black-box setting unlike the aforementioned techniques, which are focused on testing the accuracy-robustness of traditional DNNs in white-box setting


%% file: CONCLUSION.tex
\section{CONCLUSION}
\label{sec:conclusion}
In this paper, we have proposed practical black-box testing methods to evaluate energy robustness of AdNNs. The core idea behind the technique is to create inputs which increase the energy consumption of AdNN to a higher level. To achieve this goal, we have presented \name{} \footnote{https://sites.google.com/view/ereba/home}, where we have proposed two types of testing: Universal testing and Input-based testing. To our knowledge, we are the first to explore black-box testing on AdNNs. Test inputs generated by \name{} can improve the energy robustness of AdNNs. 
Finally, this paper also analyzes the behavior of AdNNs and suggests model improvement strategies. 

\section*{Acknowledgement}
This work was partially supported by Siemens Fellowship and NSF grant CCF-2146443.

%% file: main.bbl

\begin{thebibliography}{44}


\ifx \showCODEN    \undefined \def \showCODEN     #1{\unskip}     \fi
\ifx \showDOI      \undefined \def \showDOI       #1{#1}\fi
\ifx \showISBNx    \undefined \def \showISBNx     #1{\unskip}     \fi
\ifx \showISBNxiii \undefined \def \showISBNxiii  #1{\unskip}     \fi
\ifx \showISSN     \undefined \def \showISSN      #1{\unskip}     \fi
\ifx \showLCCN     \undefined \def \showLCCN      #1{\unskip}     \fi
\ifx \shownote     \undefined \def \shownote      #1{#1}          \fi
\ifx \showarticletitle \undefined \def \showarticletitle #1{#1}   \fi
\ifx \showURL      \undefined \def \showURL       {\relax}        \fi
\providecommand\bibfield[2]{#2}
\providecommand\bibinfo[2]{#2}
\providecommand\natexlab[1]{#1}
\providecommand\showeprint[2][]{arXiv:#2}

\bibitem[\protect\citeauthoryear{Benesty, Chen, Huang, and Cohen}{Benesty
  et~al\mbox{.}}{2009}]%
        {benesty2009pearson}
\bibfield{author}{\bibinfo{person}{Jacob Benesty}, \bibinfo{person}{Jingdong
  Chen}, \bibinfo{person}{Yiteng Huang}, {and} \bibinfo{person}{Israel Cohen}.}
  \bibinfo{year}{2009}\natexlab{}.
\newblock \showarticletitle{Pearson Correlation Coefficient}.
\newblock In \bibinfo{booktitle}{\emph{Noise Reduction in Speech Processing}}.
  \bibinfo{publisher}{Springer}, \bibinfo{pages}{37--40}.
\newblock


\bibitem[\protect\citeauthoryear{Bolukbasi, Wang, Dekel, and
  Saligrama}{Bolukbasi et~al\mbox{.}}{2017}]%
        {bolukbasi2017adaptive}
\bibfield{author}{\bibinfo{person}{Tolga Bolukbasi}, \bibinfo{person}{Joseph
  Wang}, \bibinfo{person}{Ofer Dekel}, {and} \bibinfo{person}{Venkatesh
  Saligrama}.} \bibinfo{year}{2017}\natexlab{}.
\newblock \showarticletitle{Adaptive Neural Networks for Efficient Inference}.
  In \bibinfo{booktitle}{\emph{Proceedings of the 34th International Conference
  on Machine Learning-Volume 70}}. JMLR. org, \bibinfo{pages}{527--536}.
\newblock


\bibitem[\protect\citeauthoryear{Bradley}{Bradley}{1997}]%
        {10.1016/S0031-3203(96)00142-2}
\bibfield{author}{\bibinfo{person}{Andrew~P. Bradley}.}
  \bibinfo{year}{1997}\natexlab{}.
\newblock \showarticletitle{The Use of the Area under the ROC Curve in the
  Evaluation of Machine Learning Algorithms}.
\newblock \bibinfo{journal}{\emph{Pattern Recogn.}} \bibinfo{volume}{30},
  \bibinfo{number}{7} (\bibinfo{date}{July} \bibinfo{year}{1997}),
  \bibinfo{pages}{1145–1159}.
\newblock
\showISSN{0031-3203}


\bibitem[\protect\citeauthoryear{Carlini and Wagner}{Carlini and
  Wagner}{2017}]%
        {carlini2017towards}
\bibfield{author}{\bibinfo{person}{Nicholas Carlini} {and}
  \bibinfo{person}{David Wagner}.} \bibinfo{year}{2017}\natexlab{}.
\newblock \showarticletitle{Towards Evaluating the Robustness of Neural
  Networks}. In \bibinfo{booktitle}{\emph{2017 IEEE Symposium on Security and
  Privacy (SP)}}. IEEE, \bibinfo{pages}{39--57}.
\newblock


\bibitem[\protect\citeauthoryear{Cheng, Dong, Pang, Su, and Zhu}{Cheng
  et~al\mbox{.}}{2019}]%
        {cheng2019improving}
\bibfield{author}{\bibinfo{person}{Shuyu Cheng}, \bibinfo{person}{Yinpeng
  Dong}, \bibinfo{person}{Tianyu Pang}, \bibinfo{person}{Hang Su}, {and}
  \bibinfo{person}{Jun Zhu}.} \bibinfo{year}{2019}\natexlab{}.
\newblock \showarticletitle{Improving Black-box Adversarial Attacks with a
  Transfer-based Prior}. In \bibinfo{booktitle}{\emph{Advances in Neural
  Information Processing Systems}}. \bibinfo{pages}{10932--10942}.
\newblock


\bibitem[\protect\citeauthoryear{Cortes and Vapnik}{Cortes and Vapnik}{1995}]%
        {cortes1995support}
\bibfield{author}{\bibinfo{person}{Corinna Cortes} {and}
  \bibinfo{person}{Vladimir Vapnik}.} \bibinfo{year}{1995}\natexlab{}.
\newblock \showarticletitle{Support-vector Networks}.
\newblock \bibinfo{journal}{\emph{Machine learning}} \bibinfo{volume}{20},
  \bibinfo{number}{3} (\bibinfo{year}{1995}), \bibinfo{pages}{273--297}.
\newblock


\bibitem[\protect\citeauthoryear{Figurnov, Collins, Zhu, Zhang, Huang, Vetrov,
  and Salakhutdinov}{Figurnov et~al\mbox{.}}{2017}]%
        {figurnov2017spatially}
\bibfield{author}{\bibinfo{person}{Michael Figurnov},
  \bibinfo{person}{Maxwell~D Collins}, \bibinfo{person}{Yukun Zhu},
  \bibinfo{person}{Li Zhang}, \bibinfo{person}{Jonathan Huang},
  \bibinfo{person}{Dmitry Vetrov}, {and} \bibinfo{person}{Ruslan
  Salakhutdinov}.} \bibinfo{year}{2017}\natexlab{}.
\newblock \showarticletitle{Spatially Adaptive Computation Time for Residual
  Networks}. In \bibinfo{booktitle}{\emph{Proceedings of the IEEE Conference on
  Computer Vision and Pattern Recognition}}. \bibinfo{pages}{1039--1048}.
\newblock


\bibitem[\protect\citeauthoryear{Gao, Zhao, Dudziak, Mullins, and Xu}{Gao
  et~al\mbox{.}}{2018}]%
        {gao2018dynamic}
\bibfield{author}{\bibinfo{person}{Xitong Gao}, \bibinfo{person}{Yiren Zhao},
  \bibinfo{person}{{\L}ukasz Dudziak}, \bibinfo{person}{Robert Mullins}, {and}
  \bibinfo{person}{Cheng-zhong Xu}.} \bibinfo{year}{2018}\natexlab{}.
\newblock \showarticletitle{Dynamic Channel Pruning: Feature Boosting and
  Suppression}.
\newblock \bibinfo{journal}{\emph{arXiv preprint arXiv:1810.05331}}
  (\bibinfo{year}{2018}).
\newblock


\bibitem[\protect\citeauthoryear{Geirhos, Rubisch, Michaelis, Bethge, Wichmann,
  and Brendel}{Geirhos et~al\mbox{.}}{2018}]%
        {geirhos2018imagenet}
\bibfield{author}{\bibinfo{person}{Robert Geirhos}, \bibinfo{person}{Patricia
  Rubisch}, \bibinfo{person}{Claudio Michaelis}, \bibinfo{person}{Matthias
  Bethge}, \bibinfo{person}{Felix~A Wichmann}, {and} \bibinfo{person}{Wieland
  Brendel}.} \bibinfo{year}{2018}\natexlab{}.
\newblock \showarticletitle{ImageNet-trained CNNs Are Biased Towards Texture;
  Increasing Shape Bias Improves Accuracy and Robustness}.
\newblock \bibinfo{journal}{\emph{arXiv preprint arXiv:1811.12231}}
  (\bibinfo{year}{2018}).
\newblock


\bibitem[\protect\citeauthoryear{Goodfellow, Shlens, and Szegedy}{Goodfellow
  et~al\mbox{.}}{2014}]%
        {goodfellow2014explaining}
\bibfield{author}{\bibinfo{person}{Ian~J Goodfellow}, \bibinfo{person}{Jonathon
  Shlens}, {and} \bibinfo{person}{Christian Szegedy}.}
  \bibinfo{year}{2014}\natexlab{}.
\newblock \showarticletitle{Explaining and Harnessing Adversarial Examples}.
\newblock \bibinfo{journal}{\emph{arXiv preprint arXiv:1412.6572}}
  (\bibinfo{year}{2014}).
\newblock


\bibitem[\protect\citeauthoryear{Graves}{Graves}{2016}]%
        {graves2016adaptive}
\bibfield{author}{\bibinfo{person}{Alex Graves}.}
  \bibinfo{year}{2016}\natexlab{}.
\newblock \showarticletitle{Adaptive Computation time for Recurrent Neural
  Networks}.
\newblock \bibinfo{journal}{\emph{arXiv preprint arXiv:1603.08983}}
  (\bibinfo{year}{2016}).
\newblock


\bibitem[\protect\citeauthoryear{Guan, Liu, Liu, and Peng}{Guan
  et~al\mbox{.}}{2017}]%
        {guan2017energy}
\bibfield{author}{\bibinfo{person}{Jiaqi Guan}, \bibinfo{person}{Yang Liu},
  \bibinfo{person}{Qiang Liu}, {and} \bibinfo{person}{Jian Peng}.}
  \bibinfo{year}{2017}\natexlab{}.
\newblock \showarticletitle{Energy-efficient Amortized Inference with Cascaded
  Deep Classifiers}.
\newblock \bibinfo{journal}{\emph{arXiv preprint arXiv:1710.03368}}
  (\bibinfo{year}{2017}).
\newblock


\bibitem[\protect\citeauthoryear{Haque, Chauhan, Liu, and Yang}{Haque
  et~al\mbox{.}}{2020}]%
        {MirazILFO}
\bibfield{author}{\bibinfo{person}{Mirazul Haque}, \bibinfo{person}{Anki
  Chauhan}, \bibinfo{person}{Cong Liu}, {and} \bibinfo{person}{Wei Yang}.}
  \bibinfo{year}{2020}\natexlab{}.
\newblock \showarticletitle{{ILFO}: Adversarial {A}ttack on {A}daptive {N}eural
  {N}etworks}. In \bibinfo{booktitle}{\emph{Proceedings of the IEEE/CVF
  Conference on Computer Vision and Pattern Recognition}}.
  \bibinfo{pages}{14264--14273}.
\newblock


\bibitem[\protect\citeauthoryear{Hendrycks and Dietterich}{Hendrycks and
  Dietterich}{2019}]%
        {hendrycks2019robustness}
\bibfield{author}{\bibinfo{person}{Dan Hendrycks} {and} \bibinfo{person}{Thomas
  Dietterich}.} \bibinfo{year}{2019}\natexlab{}.
\newblock \showarticletitle{Benchmarking Neural Network Robustness to Common
  Corruptions and Perturbations}.
\newblock \bibinfo{journal}{\emph{Proceedings of the International Conference
  on Learning Representations}} (\bibinfo{year}{2019}).
\newblock


\bibitem[\protect\citeauthoryear{Hong, Kaya, Modoranu, and Dumitra{\c{s}}}{Hong
  et~al\mbox{.}}{2020}]%
        {hong2020panda}
\bibfield{author}{\bibinfo{person}{Sanghyun Hong},
  \bibinfo{person}{Yi{\u{g}}itcan Kaya}, \bibinfo{person}{Ionu{\c{t}}-Vlad
  Modoranu}, {and} \bibinfo{person}{Tudor Dumitra{\c{s}}}.}
  \bibinfo{year}{2020}\natexlab{}.
\newblock \showarticletitle{A Panda? No, It's a Sloth: Slowdown Attacks on
  Adaptive Multi-Exit Neural Network Inference}.
\newblock \bibinfo{journal}{\emph{arXiv preprint arXiv:2010.02432}}
  (\bibinfo{year}{2020}).
\newblock


\bibitem[\protect\citeauthoryear{Hua, Zhou, De~Sa, Zhang, and Suh}{Hua
  et~al\mbox{.}}{2019}]%
        {hua2019channel}
\bibfield{author}{\bibinfo{person}{Weizhe Hua}, \bibinfo{person}{Yuan Zhou},
  \bibinfo{person}{Christopher~M De~Sa}, \bibinfo{person}{Zhiru Zhang}, {and}
  \bibinfo{person}{G~Edward Suh}.} \bibinfo{year}{2019}\natexlab{}.
\newblock \showarticletitle{Channel Gating Neural Networks}. In
  \bibinfo{booktitle}{\emph{Advances in Neural Information Processing
  Systems}}. \bibinfo{pages}{1884--1894}.
\newblock


\bibitem[\protect\citeauthoryear{Ilyas, Engstrom, Athalye, and Lin}{Ilyas
  et~al\mbox{.}}{2018}]%
        {ilyas2018black}
\bibfield{author}{\bibinfo{person}{Andrew Ilyas}, \bibinfo{person}{Logan
  Engstrom}, \bibinfo{person}{Anish Athalye}, {and} \bibinfo{person}{Jessy
  Lin}.} \bibinfo{year}{2018}\natexlab{}.
\newblock \showarticletitle{Black-box Adversarial Attacks with Limited Queries
  and Information}.
\newblock \bibinfo{journal}{\emph{arXiv preprint arXiv:1804.08598}}
  (\bibinfo{year}{2018}).
\newblock


\bibitem[\protect\citeauthoryear{K{\"o}hler, Herzog, H{\"o}nig, Wenzel, Plauth,
  Nolte, Polze, and Schr{\"o}der-Preikschat}{K{\"o}hler et~al\mbox{.}}{2020}]%
        {kohler2020pinpoint}
\bibfield{author}{\bibinfo{person}{Sven K{\"o}hler}, \bibinfo{person}{Benedict
  Herzog}, \bibinfo{person}{Timo H{\"o}nig}, \bibinfo{person}{Lukas Wenzel},
  \bibinfo{person}{Max Plauth}, \bibinfo{person}{J{\"o}rg Nolte},
  \bibinfo{person}{Andreas Polze}, {and} \bibinfo{person}{Wolfgang
  Schr{\"o}der-Preikschat}.} \bibinfo{year}{2020}\natexlab{}.
\newblock \showarticletitle{Pinpoint the Joules: Unifying Runtime-Support for
  Energy Measurements on Heterogeneous Systems}. In
  \bibinfo{booktitle}{\emph{2020 IEEE/ACM International Workshop on Runtime and
  Operating Systems for Supercomputers (ROSS)}}. IEEE, \bibinfo{pages}{31--40}.
\newblock


\bibitem[\protect\citeauthoryear{Krizhevsky}{Krizhevsky}{2009}]%
        {krizhevsky2009learning}
\bibfield{author}{\bibinfo{person}{Alex Krizhevsky}.}
  \bibinfo{year}{2009}\natexlab{}.
\newblock \showarticletitle{Learning multiple layers of features from tiny
  images}.
\newblock  (\bibinfo{year}{2009}).
\newblock


\bibitem[\protect\citeauthoryear{Liu and Deng}{Liu and Deng}{2018}]%
        {liu2018dynamic}
\bibfield{author}{\bibinfo{person}{Lanlan Liu} {and} \bibinfo{person}{Jia
  Deng}.} \bibinfo{year}{2018}\natexlab{}.
\newblock \showarticletitle{Dynamic Deep Neural Networks: Optimizing
  Accuracy-efficiency Trade-offs by Selective Execution}. In
  \bibinfo{booktitle}{\emph{Thirty-Second AAAI Conference on Artificial
  Intelligence}}.
\newblock


\bibitem[\protect\citeauthoryear{Liu, Chen, Liu, and Song}{Liu
  et~al\mbox{.}}{2016}]%
        {liu2016delving}
\bibfield{author}{\bibinfo{person}{Yanpei Liu}, \bibinfo{person}{Xinyun Chen},
  \bibinfo{person}{Chang Liu}, {and} \bibinfo{person}{Dawn Song}.}
  \bibinfo{year}{2016}\natexlab{}.
\newblock \showarticletitle{Delving into Transferable Adversarial Examples and
  Black-box Attacks}.
\newblock \bibinfo{journal}{\emph{arXiv preprint arXiv:1611.02770}}
  (\bibinfo{year}{2016}).
\newblock


\bibitem[\protect\citeauthoryear{Ma, Juefei-Xu, Zhang, Sun, Xue, Li, Chen, Su,
  Li, Liu, et~al\mbox{.}}{Ma et~al\mbox{.}}{2018a}]%
        {ma2018deepgauge}
\bibfield{author}{\bibinfo{person}{Lei Ma}, \bibinfo{person}{Felix Juefei-Xu},
  \bibinfo{person}{Fuyuan Zhang}, \bibinfo{person}{Jiyuan Sun},
  \bibinfo{person}{Minhui Xue}, \bibinfo{person}{Bo Li},
  \bibinfo{person}{Chunyang Chen}, \bibinfo{person}{Ting Su},
  \bibinfo{person}{Li Li}, \bibinfo{person}{Yang Liu}, {et~al\mbox{.}}}
  \bibinfo{year}{2018}\natexlab{a}.
\newblock \showarticletitle{Deepgauge: Multi-granularity Testing Criteria for
  Deep Learning Systems}. In \bibinfo{booktitle}{\emph{Proceedings of the 33rd
  ACM/IEEE International Conference on Automated Software Engineering}}.
  \bibinfo{pages}{120--131}.
\newblock


\bibitem[\protect\citeauthoryear{Ma, Zhang, Sun, Xue, Li, Juefei-Xu, Xie, Li,
  Liu, Zhao, et~al\mbox{.}}{Ma et~al\mbox{.}}{2018b}]%
        {ma2018deepmutation}
\bibfield{author}{\bibinfo{person}{Lei Ma}, \bibinfo{person}{Fuyuan Zhang},
  \bibinfo{person}{Jiyuan Sun}, \bibinfo{person}{Minhui Xue},
  \bibinfo{person}{Bo Li}, \bibinfo{person}{Felix Juefei-Xu},
  \bibinfo{person}{Chao Xie}, \bibinfo{person}{Li Li}, \bibinfo{person}{Yang
  Liu}, \bibinfo{person}{Jianjun Zhao}, {et~al\mbox{.}}}
  \bibinfo{year}{2018}\natexlab{b}.
\newblock \showarticletitle{Deepmutation: Mutation Testing of Deep Learning
  Systems}. In \bibinfo{booktitle}{\emph{2018 IEEE 29th International Symposium
  on Software Reliability Engineering (ISSRE)}}. IEEE,
  \bibinfo{pages}{100--111}.
\newblock


\bibitem[\protect\citeauthoryear{Nan and Saligrama}{Nan and Saligrama}{2017}]%
        {nan2017adaptive}
\bibfield{author}{\bibinfo{person}{Feng Nan} {and} \bibinfo{person}{Venkatesh
  Saligrama}.} \bibinfo{year}{2017}\natexlab{}.
\newblock \showarticletitle{Adaptive Classification for Prediction Under a
  Budget}. In \bibinfo{booktitle}{\emph{Advances in Neural Information
  Processing Systems}}. \bibinfo{pages}{4727--4737}.
\newblock


\bibitem[\protect\citeauthoryear{Nvidia}{Nvidia}{2017}]%
        {tx2}
\bibfield{author}{\bibinfo{person}{Nvidia}.} \bibinfo{year}{2017}\natexlab{}.
\newblock \bibinfo{booktitle}{\emph{{Nvidia TX2 User Manual}}}.
\newblock


\bibitem[\protect\citeauthoryear{Ovadia, Fertig, Ren, Nado, Sculley, Nowozin,
  Dillon, Lakshminarayanan, and Snoek}{Ovadia et~al\mbox{.}}{2019}]%
        {ovadia2019can}
\bibfield{author}{\bibinfo{person}{Yaniv Ovadia}, \bibinfo{person}{Emily
  Fertig}, \bibinfo{person}{Jie Ren}, \bibinfo{person}{Zachary Nado},
  \bibinfo{person}{David Sculley}, \bibinfo{person}{Sebastian Nowozin},
  \bibinfo{person}{Joshua Dillon}, \bibinfo{person}{Balaji Lakshminarayanan},
  {and} \bibinfo{person}{Jasper Snoek}.} \bibinfo{year}{2019}\natexlab{}.
\newblock \showarticletitle{Can You Trust Your Model's Uncertainty? Evaluating
  Predictive Uncertainty Under Dataset Shift}. In
  \bibinfo{booktitle}{\emph{Advances in Neural Information Processing
  Systems}}. \bibinfo{pages}{13991--14002}.
\newblock


\bibitem[\protect\citeauthoryear{Papernot, McDaniel, and Goodfellow}{Papernot
  et~al\mbox{.}}{2016a}]%
        {papernot2016transferability}
\bibfield{author}{\bibinfo{person}{Nicolas Papernot}, \bibinfo{person}{Patrick
  McDaniel}, {and} \bibinfo{person}{Ian Goodfellow}.}
  \bibinfo{year}{2016}\natexlab{a}.
\newblock \showarticletitle{Transferability in Machine Learning: from Phenomena
  to Black-box Attacks using Adversarial Samples}.
\newblock \bibinfo{journal}{\emph{arXiv preprint arXiv:1605.07277}}
  (\bibinfo{year}{2016}).
\newblock


\bibitem[\protect\citeauthoryear{Papernot, McDaniel, Goodfellow, Jha, Celik,
  and Swami}{Papernot et~al\mbox{.}}{2017}]%
        {papernot2017practical}
\bibfield{author}{\bibinfo{person}{Nicolas Papernot}, \bibinfo{person}{Patrick
  McDaniel}, \bibinfo{person}{Ian Goodfellow}, \bibinfo{person}{Somesh Jha},
  \bibinfo{person}{Z~Berkay Celik}, {and} \bibinfo{person}{Ananthram Swami}.}
  \bibinfo{year}{2017}\natexlab{}.
\newblock \showarticletitle{Practical Black-box Attacks against Machine
  Learning}. In \bibinfo{booktitle}{\emph{Proceedings of the 2017 ACM on Asia
  conference on computer and communications security}}.
  \bibinfo{pages}{506--519}.
\newblock


\bibitem[\protect\citeauthoryear{Papernot, McDaniel, Jha, Fredrikson, Celik,
  and Swami}{Papernot et~al\mbox{.}}{2016b}]%
        {papernot2016limitations}
\bibfield{author}{\bibinfo{person}{Nicolas Papernot}, \bibinfo{person}{Patrick
  McDaniel}, \bibinfo{person}{Somesh Jha}, \bibinfo{person}{Matt Fredrikson},
  \bibinfo{person}{Z~Berkay Celik}, {and} \bibinfo{person}{Ananthram Swami}.}
  \bibinfo{year}{2016}\natexlab{b}.
\newblock \showarticletitle{The Limitations of Deep Learning in Adversarial
  Settings}. In \bibinfo{booktitle}{\emph{2016 IEEE European Symposium on
  Security and Privacy (EuroS\&P)}}. IEEE, \bibinfo{pages}{372--387}.
\newblock


\bibitem[\protect\citeauthoryear{Pei, Cao, Yang, and Jana}{Pei
  et~al\mbox{.}}{2017}]%
        {pei2017deepxplore}
\bibfield{author}{\bibinfo{person}{Kexin Pei}, \bibinfo{person}{Yinzhi Cao},
  \bibinfo{person}{Junfeng Yang}, {and} \bibinfo{person}{Suman Jana}.}
  \bibinfo{year}{2017}\natexlab{}.
\newblock \showarticletitle{Deepxplore: Automated Whitebox Testing of Deep
  Learning Systems}. In \bibinfo{booktitle}{\emph{proceedings of the 26th
  Symposium on Operating Systems Principles}}. \bibinfo{pages}{1--18}.
\newblock


\bibitem[\protect\citeauthoryear{Quionero-Candela, Sugiyama, Schwaighofer, and
  Lawrence}{Quionero-Candela et~al\mbox{.}}{2009}]%
        {quionero2009dataset}
\bibfield{author}{\bibinfo{person}{Joaquin Quionero-Candela},
  \bibinfo{person}{Masashi Sugiyama}, \bibinfo{person}{Anton Schwaighofer},
  {and} \bibinfo{person}{Neil~D Lawrence}.} \bibinfo{year}{2009}\natexlab{}.
\newblock \bibinfo{booktitle}{\emph{Dataset shift in machine learning}}.
\newblock \bibinfo{publisher}{The MIT Press}.
\newblock


\bibitem[\protect\citeauthoryear{Szegedy, Zaremba, Sutskever, Bruna, Erhan,
  Goodfellow, and Fergus}{Szegedy et~al\mbox{.}}{2013}]%
        {szegedy2013intriguing}
\bibfield{author}{\bibinfo{person}{Christian Szegedy},
  \bibinfo{person}{Wojciech Zaremba}, \bibinfo{person}{Ilya Sutskever},
  \bibinfo{person}{Joan Bruna}, \bibinfo{person}{Dumitru Erhan},
  \bibinfo{person}{Ian Goodfellow}, {and} \bibinfo{person}{Rob Fergus}.}
  \bibinfo{year}{2013}\natexlab{}.
\newblock \showarticletitle{Intriguing Properties of Neural Networks}.
\newblock \bibinfo{journal}{\emph{arXiv preprint arXiv:1312.6199}}
  (\bibinfo{year}{2013}).
\newblock


\bibitem[\protect\citeauthoryear{Teerapittayanon, McDanel, and
  Kung}{Teerapittayanon et~al\mbox{.}}{2016}]%
        {teerapittayanon2016branchynet}
\bibfield{author}{\bibinfo{person}{Surat Teerapittayanon},
  \bibinfo{person}{Bradley McDanel}, {and} \bibinfo{person}{Hsiang-Tsung
  Kung}.} \bibinfo{year}{2016}\natexlab{}.
\newblock \showarticletitle{Branchynet: Fast Inference via Early Exiting from
  Deep Neural Networks}. In \bibinfo{booktitle}{\emph{2016 23rd International
  Conference on Pattern Recognition (ICPR)}}. IEEE,
  \bibinfo{pages}{2464--2469}.
\newblock


\bibitem[\protect\citeauthoryear{Teerapittayanon, McDanel, and
  Kung}{Teerapittayanon et~al\mbox{.}}{2017}]%
        {teerapittayanon2017distributed}
\bibfield{author}{\bibinfo{person}{Surat Teerapittayanon},
  \bibinfo{person}{Bradley McDanel}, {and} \bibinfo{person}{Hsiang-Tsung
  Kung}.} \bibinfo{year}{2017}\natexlab{}.
\newblock \showarticletitle{Distributed Deep Neural Networks over the Cloud,
  the Edge and End Devices}. In \bibinfo{booktitle}{\emph{2017 IEEE 37th
  International Conference on Distributed Computing Systems (ICDCS)}}. IEEE,
  \bibinfo{pages}{328--339}.
\newblock


\bibitem[\protect\citeauthoryear{Tensorflow}{Tensorflow}{2009a}]%
        {cifar10}
\bibfield{author}{\bibinfo{person}{Tensorflow}.}
  \bibinfo{year}{2009}\natexlab{a}.
\newblock \bibinfo{title}{{Tensorflow Deep Learning Framework.}}
\newblock
  \bibinfo{howpublished}{\url{https://www.tensorflow.org/datasets/catalog/cifar10}}.
\newblock


\bibitem[\protect\citeauthoryear{Tensorflow}{Tensorflow}{2009b}]%
        {cifar100}
\bibfield{author}{\bibinfo{person}{Tensorflow}.}
  \bibinfo{year}{2009}\natexlab{b}.
\newblock \bibinfo{title}{{Tensorflow Deep Learning Framework.}}
\newblock
  \bibinfo{howpublished}{\url{https://www.tensorflow.org/datasets/catalog/cifar100}}.
\newblock


\bibitem[\protect\citeauthoryear{Tian, Pei, Jana, and Ray}{Tian
  et~al\mbox{.}}{2018}]%
        {tian2018deeptest}
\bibfield{author}{\bibinfo{person}{Yuchi Tian}, \bibinfo{person}{Kexin Pei},
  \bibinfo{person}{Suman Jana}, {and} \bibinfo{person}{Baishakhi Ray}.}
  \bibinfo{year}{2018}\natexlab{}.
\newblock \showarticletitle{Deeptest: Automated Testing of
  Deep-neural-network-driven Autonomous Cars}. In
  \bibinfo{booktitle}{\emph{Proceedings of the 40th international conference on
  software engineering}}. \bibinfo{pages}{303--314}.
\newblock


\bibitem[\protect\citeauthoryear{Wang, Yu, Dou, Darrell, and Gonzalez}{Wang
  et~al\mbox{.}}{2018}]%
        {wang2018skipnet}
\bibfield{author}{\bibinfo{person}{Xin Wang}, \bibinfo{person}{Fisher Yu},
  \bibinfo{person}{Zi-Yi Dou}, \bibinfo{person}{Trevor Darrell}, {and}
  \bibinfo{person}{Joseph~E Gonzalez}.} \bibinfo{year}{2018}\natexlab{}.
\newblock \showarticletitle{Skipnet: Learning Dynamic Routing in Convolutional
  Networks}. In \bibinfo{booktitle}{\emph{Proceedings of the European
  Conference on Computer Vision (ECCV)}}. \bibinfo{pages}{409--424}.
\newblock


\bibitem[\protect\citeauthoryear{Wikipedia}{Wikipedia}{[n.d.]a}]%
        {psnr}
\bibfield{author}{\bibinfo{person}{Wikipedia}.}
  \bibinfo{year}{[n.d.]}\natexlab{a}.
\newblock \bibinfo{title}{{Peak Signal-to-Noise Ratio}}.
\newblock
  \bibinfo{howpublished}{\url{https://en.wikipedia.org/wiki/Peak_signal-to-noise_ratio}}.
\newblock


\bibitem[\protect\citeauthoryear{Wikipedia}{Wikipedia}{[n.d.]b}]%
        {SSIM}
\bibfield{author}{\bibinfo{person}{Wikipedia}.}
  \bibinfo{year}{[n.d.]}\natexlab{b}.
\newblock \bibinfo{title}{{Structural Similarity Index Measure}}.
\newblock
  \bibinfo{howpublished}{\url{https://en.wikipedia.org/wiki/Structural_similarity}}.
\newblock


\bibitem[\protect\citeauthoryear{Wu, Nagarajan, Kumar, Rennie, Davis, Grauman,
  and Feris}{Wu et~al\mbox{.}}{2018}]%
        {wu2018blockdrop}
\bibfield{author}{\bibinfo{person}{Zuxuan Wu}, \bibinfo{person}{Tushar
  Nagarajan}, \bibinfo{person}{Abhishek Kumar}, \bibinfo{person}{Steven
  Rennie}, \bibinfo{person}{Larry~S Davis}, \bibinfo{person}{Kristen Grauman},
  {and} \bibinfo{person}{Rogerio Feris}.} \bibinfo{year}{2018}\natexlab{}.
\newblock \showarticletitle{Blockdrop: Dynamic Inference Paths in Residual
  Networks}. In \bibinfo{booktitle}{\emph{Proceedings of the IEEE Conference on
  Computer Vision and Pattern Recognition}}. \bibinfo{pages}{8817--8826}.
\newblock


\bibitem[\protect\citeauthoryear{Xie, Luong, Hovy, and Le}{Xie
  et~al\mbox{.}}{2020}]%
        {xie2020self}
\bibfield{author}{\bibinfo{person}{Qizhe Xie}, \bibinfo{person}{Minh-Thang
  Luong}, \bibinfo{person}{Eduard Hovy}, {and} \bibinfo{person}{Quoc~V Le}.}
  \bibinfo{year}{2020}\natexlab{}.
\newblock \showarticletitle{Self-training with Noisy Student Improves Imagenet
  Classification}. In \bibinfo{booktitle}{\emph{Proceedings of the IEEE/CVF
  Conference on Computer Vision and Pattern Recognition}}.
  \bibinfo{pages}{10687--10698}.
\newblock


\bibitem[\protect\citeauthoryear{Xie, Ma, Juefei-Xu, Xue, Chen, Liu, Zhao, Li,
  Yin, and See}{Xie et~al\mbox{.}}{2019}]%
        {xie2019deephunter}
\bibfield{author}{\bibinfo{person}{Xiaofei Xie}, \bibinfo{person}{Lei Ma},
  \bibinfo{person}{Felix Juefei-Xu}, \bibinfo{person}{Minhui Xue},
  \bibinfo{person}{Hongxu Chen}, \bibinfo{person}{Yang Liu},
  \bibinfo{person}{Jianjun Zhao}, \bibinfo{person}{Bo Li},
  \bibinfo{person}{Jianxiong Yin}, {and} \bibinfo{person}{Simon See}.}
  \bibinfo{year}{2019}\natexlab{}.
\newblock \showarticletitle{DeepHunter: a Coverage-guided Fuzz Testing
  Framework for Deep Neural Networks}. In \bibinfo{booktitle}{\emph{Proceedings
  of the 28th ACM SIGSOFT International Symposium on Software Testing and
  Analysis}}. \bibinfo{pages}{146--157}.
\newblock


\bibitem[\protect\citeauthoryear{Yang, Han, Chen, Song, Dai, and Huang}{Yang
  et~al\mbox{.}}{2020}]%
        {yang2020resolution}
\bibfield{author}{\bibinfo{person}{Le Yang}, \bibinfo{person}{Yizeng Han},
  \bibinfo{person}{Xi Chen}, \bibinfo{person}{Shiji Song},
  \bibinfo{person}{Jifeng Dai}, {and} \bibinfo{person}{Gao Huang}.}
  \bibinfo{year}{2020}\natexlab{}.
\newblock \showarticletitle{Resolution Adaptive Networks for Efficient
  Inference}. In \bibinfo{booktitle}{\emph{Proceedings of the IEEE Conference
  on Computer Vision and Pattern Recognition}}. \bibinfo{pages}{2366--2375}.
\newblock


\end{thebibliography}
